\documentclass[11pt]{article}

\usepackage{acl}
\usepackage{times}
\usepackage{latexsym}
\usepackage[T1]{fontenc}
\usepackage[utf8]{inputenc}
\usepackage{microtype}
\usepackage{inconsolata}
\usepackage{fancyvrb}
\usepackage{graphicx}
\usepackage{amsmath}
\usepackage{amssymb}
\usepackage{booktabs}
\usepackage{multirow}
\usepackage{caption}
\usepackage{subcaption}

\title{When Do Prompt-Side Agent Playbooks Transfer?\\
Accuracy, Cost, and Runtime Shift in Agent Deployment}

\author{Weihong Lin \quad Lin Sun\thanks{Corresponding author.} \quad Xiangzheng Zhang \\
  Beijing Qiyuan Technology Co., Ltd., Beijing, China \\
  \texttt{sunlin1@360.cn}}

\begin{document}
\maketitle
\begin{abstract}
Prompt-side playbooks can improve tool-using language agents without retraining, but their portability beyond the source setting is unclear. We study frozen playbook transfer under a shared distill--validate--transfer protocol. On ALFWorld, transfer is beneficial under controlled greedy decoding and, in one near-budget-matched comparison, distilled guidance outperforms five fixed demonstrations. On TAU2-Bench, a prespecified aggregate contrast supports a modest average matched-domain advantage, but global Holm correction retains only one of 135 route-level effects; the remaining grid provides descriptive evidence of compatibility-sensitive heterogeneity. On XBench-DeepSearch, one artifact--runtime pairing preserves useful first-try heuristics while producing repeated queries, delayed stopping, and substantial cost inflation after a context-runtime shift. Across benchmarks, transferred and target-derived playbooks both require target-side validation of success, termination, protocol compatibility, and cost. Frozen transfer is therefore a conditional cold-start option, not a reuse-by-default strategy or a universally preferable alternative to target-side redistillation.
\end{abstract}

\section{Introduction}

Prompt-side playbooks are becoming a common way to improve agent systems in production. After repeated failures and recoveries, teams summarize useful procedures into compact instructions: what to check first, which tools to prioritize, when to recover, and how to avoid common failure modes. These playbooks are attractive because they can be deployed immediately and reused across related systems without retraining.

That convenience creates a deployment question: when does a playbook learned in one setting remain useful after transfer to another? A playbook can be moved across target domains, model scales, model families, or runtime regimes. In practice, the same transferred procedure may help one target, do little on another, and harm a third. The cleanest positive case is real: on ALFWorld, one transferred playbook helps eight greedy-decoding targets and usually shortens trajectories by rescuing runs that would otherwise hit the 50-step limit. But that positive case does not generalize by default. In more heterogeneous service-agent settings the same kind of artifact becomes compatibility-sensitive, and under runtime shift its stopping behavior can become misaligned.

This paper studies frozen playbook transfer: compact prompt-side procedures distilled from prior runs, validated on the source side, and then transferred without target-side retuning. Frozen reuse is not our preferred default. It is a cold-start or amortized candidate when target-side failure trajectories are not yet available, when collecting target traces creates privacy, cost, or latency constraints, or when an existing artifact can be screened before investing in target-side construction. Once target evidence is available, redistilling, repairing, or rejecting the artifact may be preferable. We do not propose a new learning algorithm; we ask when a frozen candidate remains useful after transfer.

Our main finding is simple: prompt-side playbooks can transfer, but only conditionally. Across three benchmarks, the results are most consistent with an operational-compatibility view: transfer works when the assumptions encoded in a playbook remain compatible with the target deployment setting. Those assumptions include workflow, target capability, decoding regime, and runtime budget. When they remain aligned, transfer can improve success and sometimes reduce trajectory length. When they do not, the same playbook can inject the wrong action pattern, waste turns, or destabilize stopping behavior.

We evaluate transfer under a shared \emph{distill--validate--transfer} protocol on three benchmarks that play different roles in the argument. ALFWorld \citep{shridhar2021alfworld} provides a controlled existence test: can a frozen playbook transfer at all? TAU2-Bench \citep{tau2bench2024} is our main service-agent benchmark: does transfer survive realistic domain and model heterogeneity? XBench-DeepSearch adds a runtime-shift stress test: does a transferred playbook remain operationally stable when the runtime regime changes?

Our contributions are threefold:
\begin{itemize}
\setlength{\itemsep}{0pt}
\setlength{\topsep}{2pt}
\setlength{\parsep}{0pt}
\setlength{\partopsep}{0pt}
\item We define a deployment-oriented protocol for evaluating frozen prompt-side playbook transfer rather than target-side retuning.
\item We show that transfer is real but not uniform: positive cases appear under controlled conditions, while a multiplicity-corrected service-agent analysis supports a modest matched-domain advantage but little route-level confirmatory evidence.
\item We compare distilled guidance with near-budget-matched fixed demonstrations and compare frozen transfer with target-side redistillation, clarifying what frozen reuse does and does not offer.
\item We show that deployment evaluation must include operational cost as well as task success, because an artifact--runtime mismatch can preserve useful heuristics while destabilizing stopping.
\end{itemize}

The practical takeaway is straightforward: teams should treat playbook reuse as a validation problem, not as a reuse-by-default strategy.

\section{Related Work}

Prior work improves agents through reflection, self-feedback, experience replay, skill distillation, and prompt optimization \citep{shinn2023reflexion,madaan2023selfrefine,zhao2024expel,liu2025cer,shalev2026training,ni2026trace2skill,wang2024voyager,xu2025amem,ravindran2026portable,wei2022chain,yao2023react,wang2024promptagent,yang2024opro}, showing that agent experience can be externalized and reused. Our focus is narrower and more deployment-oriented: rather than proposing another improvement mechanism, we ask whether an already-derived artifact remains useful after transfer across target models, domains, families, or runtime conditions. We add a fixed few-shot comparator, but human-authored playbooks and deployment-time retrieval systems answer broader method-selection questions and remain outside our empirical comparison. Recalled experience can be over-followed and may propagate stale lessons \citep{xiong2025memory}, making negative transfer a realistic operational failure mode.

\section{Evaluating Frozen Playbook Transfer}

\subsection{What Is Transferred and How}

We study a playbook: a frozen prompt-side operating procedure distilled from repeated source-side executions. It is not a retrained policy or a full memory system; it is a compact operational artifact inserted into the agent prompt at deployment.

\begin{figure}[t]
\centering
\includegraphics[width=0.94\columnwidth]{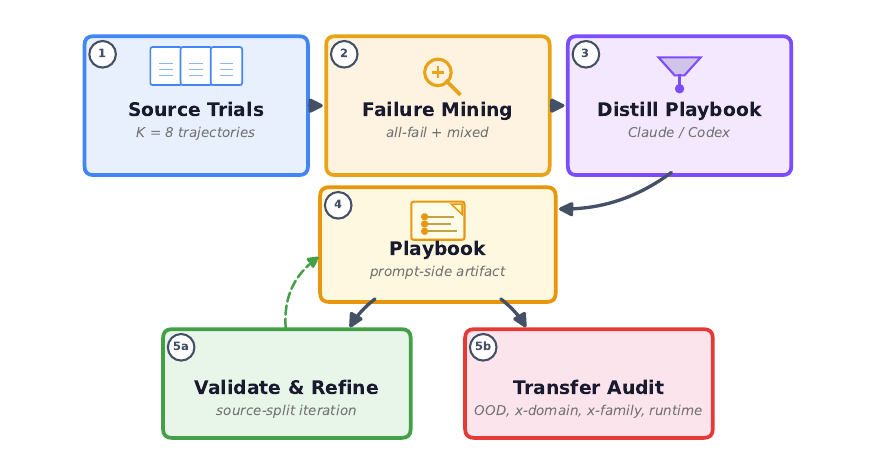}
\caption{Shared distill--validate--transfer protocol. A stronger teacher distills a playbook from repeated source trajectories; the final artifact is validated on the source split and frozen for transfer.}
\label{fig:method_pipeline}
\end{figure}

Figure~\ref{fig:method_pipeline} summarizes the shared \emph{distill--validate--transfer} protocol. We repeatedly run a source agent, mine \emph{all-fail} and \emph{mixed-outcome} bundles, and give them to a stronger closed-source distiller: Claude Opus 4.6 via Claude Code or GPT-5.4 via Codex, depending on the artifact family. The distiller abstracts recurrent rules, search priorities, and recovery heuristics into a compact playbook, validates it on the source side, and we then freeze it before transfer. We therefore study portability of a \emph{frozen prompt-side artifact}, not target-side prompt retuning. The main controls are no-playbook baselines, same-length placebos, and matched-versus-mismatched transfer conditions.

\subsection{Evaluation Axes and Benchmarks}

We evaluate transfer along three deployment axes: task setting (scene or domain), target model (size or family), and runtime regime (context budget or decoding). Across all settings, we track both task success and operational cost.

The three benchmarks play complementary roles. ALFWorld \citep{shridhar2021alfworld} is the controlled existence test, covering scene OOD, target scale, and cross-family transfer under matched decoding. TAU2-Bench \citep{tau2bench2024} is the main service-agent benchmark, spanning \texttt{airline}, \texttt{retail}, and \texttt{telecom} under both in-family and cross-family transfer. XBench-DeepSearch is a small runtime-shift stress test: it probes 32K-era playbooks under later 128K execution settings. Benchmark-specific audits and transfer tables are reported in Appendix Sections~\ref{sec:app_alfworld}, \ref{sec:app_tau2}, and~\ref{sec:app_xbench}.

\subsection{Artifacts and Metrics}
\label{sec:artifact_naming}

For each source condition, the transferred object is the final playbook produced by the protocol above. On ALFWorld, we study four artifacts: C7S/C32S are Claude-distilled playbooks, while X7D/X7C are Codex-distilled detailed/compact variants from the same 7B source pool. On TAU2-Bench, source playbooks are named by source size and domain, e.g., P14R = 14B-retail; the 7B/14B/32B size shorthands refer to Qwen2.5-Instruct checkpoints. On XBench-DeepSearch, we study one Claude-distilled XBench playbook derived from Tongyi-DeepResearch-30B-A3B 32K source-side trajectories and then transfer it across model and context conditions. Tables~\ref{tab:app_artifact_alf} and~\ref{tab:app_artifact_tau2} give the detailed ALFWorld and TAU2-Bench inventories, and Appendix~\ref{app:artifact_inventory} also provides one complete example playbook (P14R) plus a short cross-stack excerpt (X7D).

On ALFWorld, we report normalized success rate with environment-broken games removed from the denominator. On TAU2-Bench, we use corrected official-style Pass$^k$, with Pass$^1$ as the sole route-level headline metric. We apply global Holm correction to its 135 route hypotheses; Pass$^{2\text{--}4}$, cost, and unretained routes remain descriptive. On XBench-DeepSearch, we report Pass@1, question-keyed True Pass@3, and deployment-side cost metrics. Full lift tables, CI audits, and descriptive transfer tables are deferred to Appendix Sections~\ref{sec:app_alfworld}, \ref{sec:app_tau2}, and~\ref{sec:app_xbench}.

\section{Controlled Transfer Can Work (ALFWorld)}

We begin with ALFWorld because it gives the cleanest answer to the first paper-level question: can prompt-side experience transfer work at all before we enter a more heterogeneous service-agent setting? On ALFWorld, the answer is yes, but the 400-run audit sharpens that result: portability is also decoding-dependent.

\subsection{Transfer Under Controlled Decoding}

\begin{figure*}[t]
\centering
\includegraphics[width=0.47\textwidth]{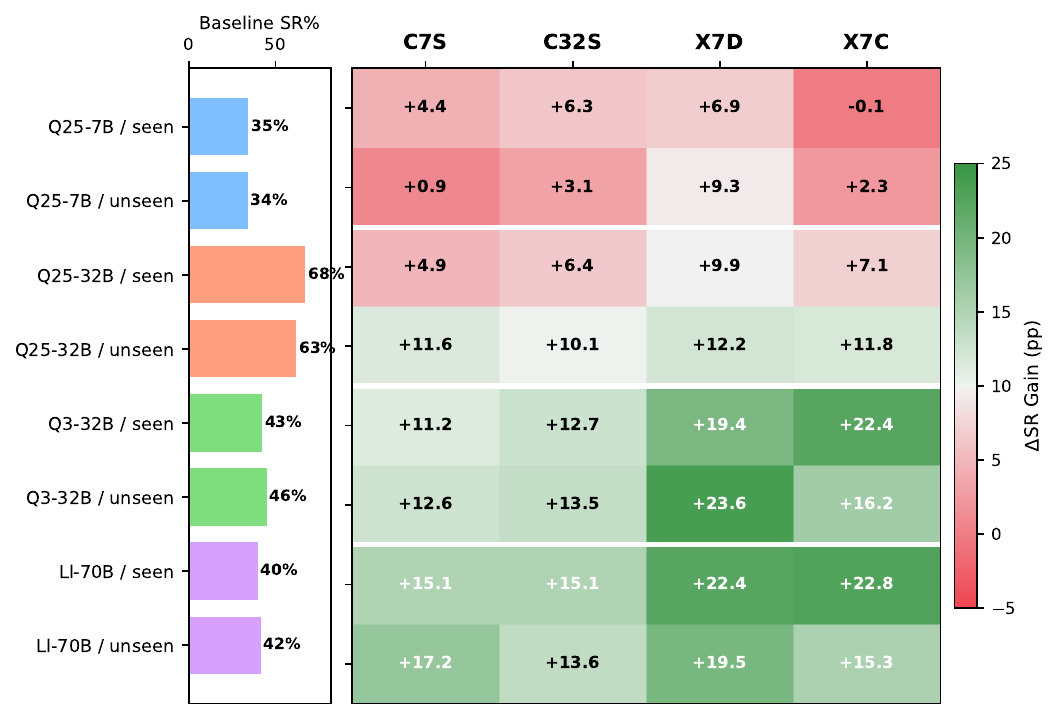}%
\hfill
\includegraphics[width=0.47\textwidth]{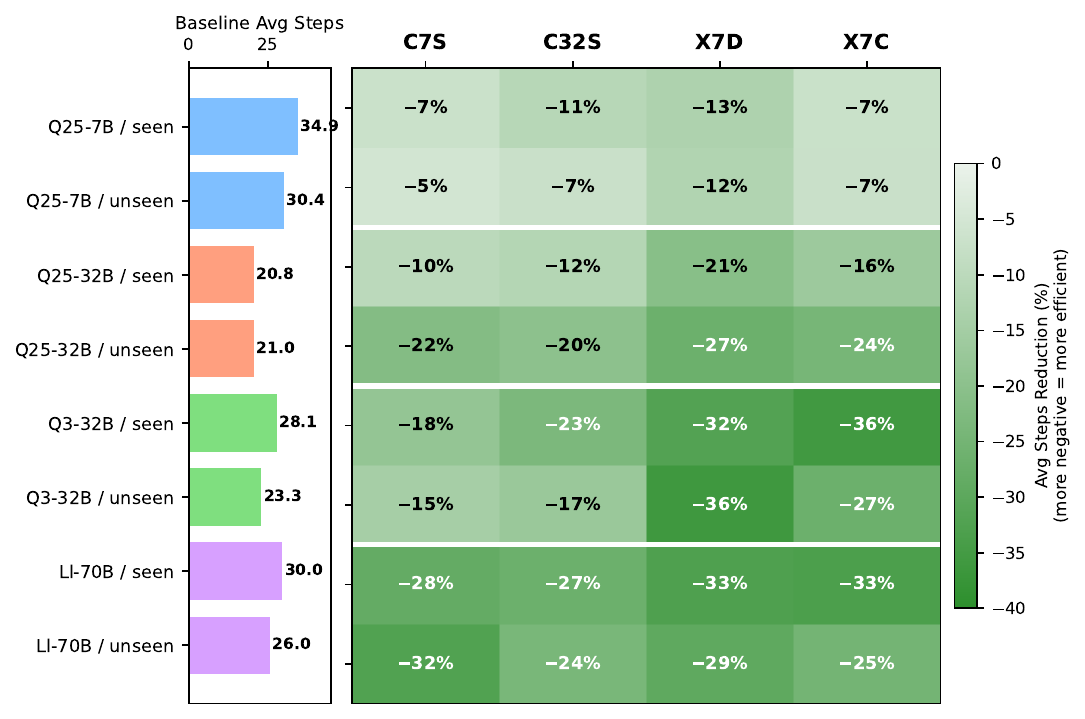}
\caption{ALFWorld transfer at $t{=}0$. Left: $\Delta$SR. Right: avg interaction-step reduction. Greedy transfer is strongest on larger targets and comes with shorter trajectories.}
\label{fig:alf_combined}
\end{figure*}

Under greedy decoding ($t{=}0$), transferred playbooks help across scene-OOD evaluation, target scale, and model family. The effect grows with target capacity: gains are modest on Qwen2.5-7B-Instruct, uniformly positive on Qwen2.5-32B-Instruct, and especially strong on Qwen3-32B and Llama3.1-70B-Inst, where the best lift reaches +23.6 pp.
Figure~\ref{fig:alf_combined} summarizes this controlled-decoding picture: larger targets receive the strongest success gains, and those gains are accompanied by shorter interaction traces rather than cost inflation.

\subsection{Decoding Robustness Depends on Capacity and Artifact Specificity}

\begin{figure}[t]
\centering
\includegraphics[width=0.94\columnwidth]{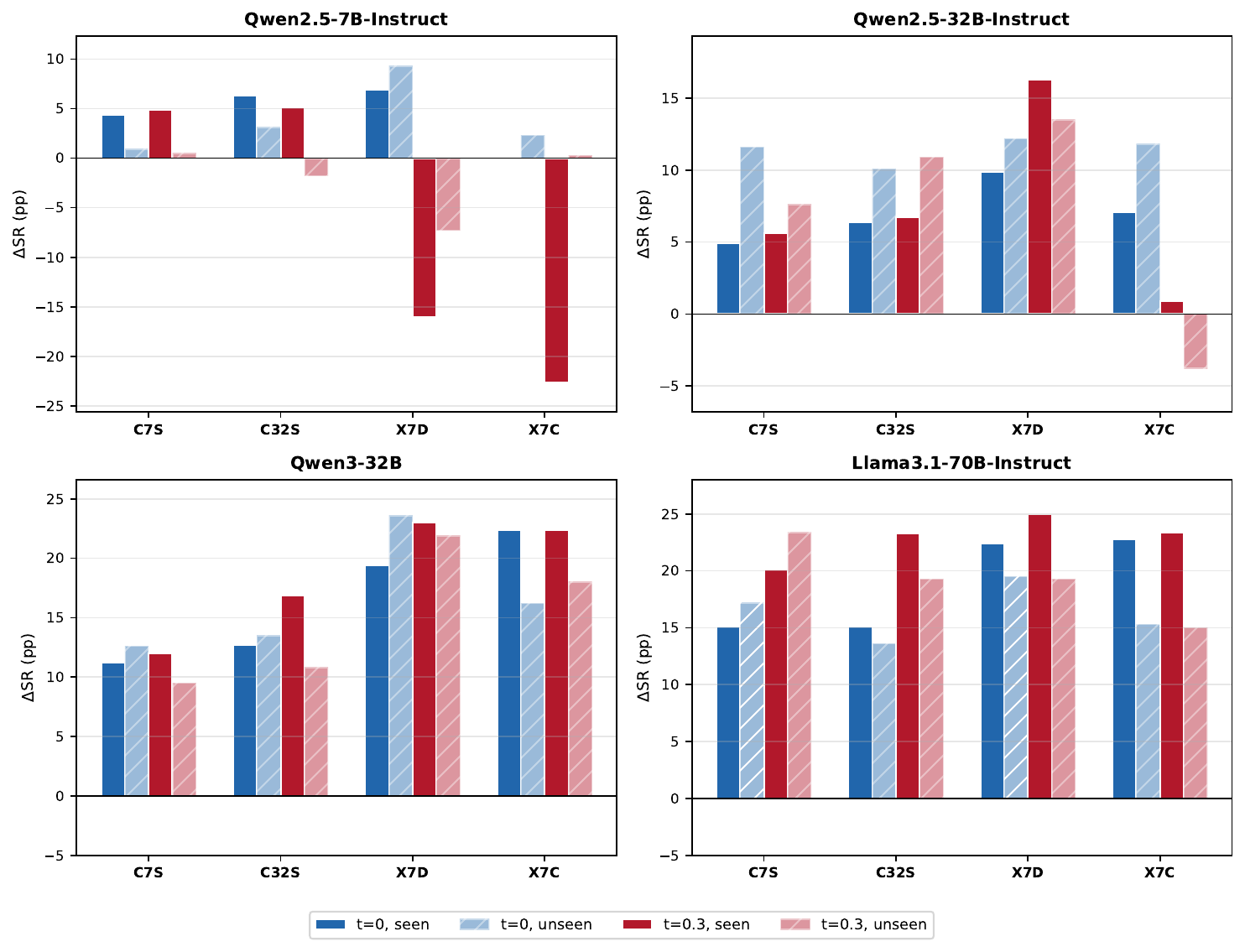}
\caption{Temperature robustness on ALFWorld. Bars show $\Delta$SR vs.\ baseline at $t{=}0$ and $t{=}0.3$ on \texttt{seen} and \texttt{unseen}.}
\label{fig:temp_robustness}
\end{figure}

The matched $t{=}0$ vs.\ $t{=}0.3$ audit in Figure~\ref{fig:temp_robustness} then shows why ALFWorld remains valuable even after that existence proof. The temperature effect is not random noise; it interacts sharply with target capacity and playbook form. On Qwen2.5-7B-Instruct, the Codex-7B-detailed artifact (X7D) flips from positive at $t{=}0$ to clearly harmful at $t{=}0.3$, and the cost metrics move with it: hit-50 rises from 58.1\% to 79.8\% on \texttt{seen}, while mean trajectory length jumps from 32.8 to 42.4 steps.

On Qwen2.5-32B-Instruct, X7D stays robust while the shorter X7C collapses sharply. Qwen3-32B and Llama3.1-70B-Inst show the opposite pattern: sampling improves the baseline and keeps all four playbooks positive. ALFWorld therefore contributes a controlled existence proof and robustness audit: transfer \emph{can} work under controlled decoding, but robustness depends on decoding regime $\times$ target capacity $\times$ artifact specificity. Appendix~\ref{sec:app_alfworld} reports the full 400-run audit.

\subsection{Distilled Guidance versus Fixed Demonstrations}

We compare X7D with a five-example fixed few-shot prompt (FS-5) on the Qwen3-32B \texttt{valid\_unseen} target. Both artifacts use the same Qwen2.5-7B \texttt{valid\_train} source run. A source-only rule selects five canonical successes for FS-5, and both prompts and the 148-game manifest are frozen before evaluation. Under near-matched artifact budgets (4,191 versus 4,001 bytes), X7D succeeds on 107/148 games (72.3\%) versus 91/148 (61.5\%) for FS-5: $+10.8$ pp, CI$_{95}$ [$+2.7$,$+18.9$], exact McNemar $p{=}.0139$. This establishes incremental value over fixed demonstrations in one compatible transfer setting; it does not establish superiority over human-authored guidance or dynamic retrieval.

\section{Transfer Is Compatibility-Sensitive (TAU2-Bench)}

TAU2-Bench is the paper's main service-agent result because it matches deployment heterogeneity across business domains, target scales, and model families. Its strongest design-level inference is a modest average matched-domain advantage; route-level effects are mostly descriptive after correction for multiplicity. We therefore use \emph{compatibility-sensitive transfer} to describe aggregate domain alignment together with heterogeneous configuration-level point estimates, not broadly detectable configuration-specific gains.

\subsection{In-Family Transfer Landscape}

\begin{figure*}[t]
\centering
\includegraphics[width=0.92\textwidth]{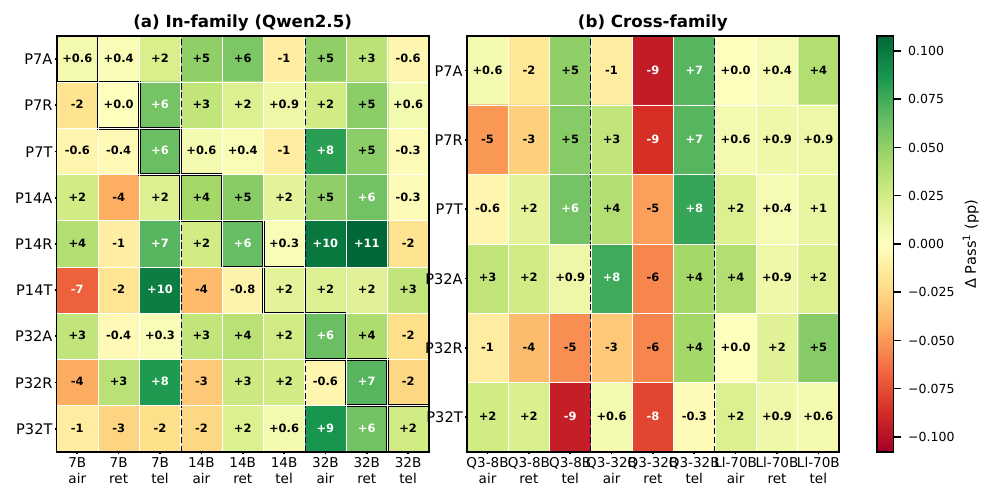}
\caption{Descriptive TAU2-Bench $\Delta$Pass$^1$ heatmaps for in-family and cross-family transfer. Full Pass$^{1\text{--}4}$ heatmaps and paired intervals are in Appendix~\ref{sec:app_tau2_heatmaps}; only the routes in Table~\ref{tab:tau2_multiplicity} survive a reported multiplicity correction.}
\label{fig:tau2_pass1_heatmap}
\end{figure*}

The full Qwen2.5-Instruct in-family grid in Figure~\ref{fig:tau2_pass1_heatmap} covers 81 source--target--domain configurations. Cell-wise intervals are often wide, and only one in-family route survives Holm correction over this family. We therefore use the heatmap to describe direction and magnitude: its point estimates are non-uniform, with visible positive regions in 32B/retail, 32B/airline, and 7B/telecom, but these regions are not individually confirmatory findings.

The same grid provides a control against a pure prompt-injection explanation. For each of nine source artifacts, we compare the \emph{same injected file} on its matched domain and two mismatched domains, controlling prompt length and format within source. This separately defined, route-agnostic contrast uses all 81 eligible in-family configurations and is not selected from the 135 route hypotheses. Matched-domain transfer is stronger by $+1.72$ pp on average (CI$_{95}$ [$+0.25$,$+2.97$], exact blocked-permutation $p{=}.035$; Appendix~\ref{sec:app_tau2_domain_control}). It supports a modest average domain-alignment advantage, not improvement on most routes.

\begin{table}[t]
\centering
\scriptsize
\setlength{\tabcolsep}{3.0pt}
\begin{tabular}{@{}llrrr@{}}
\toprule
Route & $\Delta$Pass$^1$ & Fam. Holm & Global Holm & Global BH \\
\midrule
P14T$\to$Q2.5-7B/T & $+9.69$ & .0196 & .0327 & .0316 \\
P32T$\to$Q3-8B/T & $-9.38$ & .0299 & .0742 & .0316 \\
P7A$\to$Q3-32B/R & $-9.48$ & .0372 & .0934 & .0316 \\
\bottomrule
\end{tabular}
\caption{TAU2 Pass$^1$ routes retained by either family-specific Holm or global BH correction. Global Holm over all 135 routes retains only the first route. BH is an FDR sensitivity analysis; Pass$^{2\text{--}4}$ and cost remain descriptive.}
\label{tab:tau2_multiplicity}
\end{table}

This is the main interpretive shift relative to ALFWorld. TAU2-Bench does not support a broad ``playbooks help'' reading. Global Holm correction over Pass$^1$ retains 1/135 positive route; family-specific Holm additionally retains two cross-family negative routes, which also appear in a global Benjamini--Hochberg sensitivity analysis. The aggregate contrast is therefore the primary TAU2 inference, while the grid describes compatibility-sensitive heterogeneity. Failure to survive correction is not interpreted as evidence of zero effect.

\subsection{Cross-Family Transfer Landscape}

The cross-family point estimates push this conditionality further. They do not fit a monotonic ``farther family, weaker effect'' story; instead, signs vary by target and domain. After family-specific Holm correction, two negative routes remain; neither survives global Holm, while both remain in the global BH sensitivity analysis. Other apparent gains and losses are descriptive.

Appendix Figure~\ref{fig:tau2_summary_combined} summarizes the target-condition view. The in-family baseline-versus-transfer slope is weak, while the cross-family slope turns negative; this remains a descriptive pattern rather than route-level inference.

\subsection{Illustrative Configurations and Content Placebo}

Appendix Table~\ref{tab:app_tau2_supported} reports the multiplicity-retained routes, while the following cases illustrate possible behavioral pathways without adding inferential evidence. The full P14T$\to$7B/airline route changes by approximately $-6.9$ pp and does not survive global correction. Within it, one deliberately selected task drops from 62\% to 12\% when an imported telecom complaint template injects an unnecessary \texttt{send\_certificate} action. Conversely, a selected P14R$\to$14B/retail task rises from 56\% to 88\% after the playbook inserts a missing prerequisite read. Neither task-level example is representative of the full grid.

To distinguish content-driven transfer from format-driven effects, we add a same-length generic placebo on one representative positive route and two representative negative routes, including a repeated cross-family failure condition. Each placebo is a structure-matched rewrite of its source playbook into generic workflow advice, generated with Claude Opus 4.6 via Claude Code: it keeps comparable length, sectioning, and checklist style, but removes domain-specific procedures, tool names, prerequisite logic, and action-ordering rules.

\begin{table}[t]
\centering
\small
\setlength{\tabcolsep}{4pt}
\begin{tabular}{@{}lccc@{}}
\toprule
Route & Baseline & Playbook & Placebo \\
\midrule
P14R$\to$14B/R & 55.2 & 62.6 & 57.1 \\
P14T$\to$7B/A  & 15.6 & 8.7 & 14.4 \\
P7A$\to$Q3-32B/R & 72.4 & 60.8 & 70.7 \\
\bottomrule
\end{tabular}
\caption{Descriptive same-length prompt control on three selected TAU2 routes (Pass$^1$, \%). The placebo remains near baseline; this is a small content control, not a benchmark-wide format ablation.}
\label{tab:tau2_placebo_main}
\end{table}

Across these three routes, the same-length placebo stays near baseline while the transferred playbook reproduces the corresponding point-estimate direction, weakening a pure prompt-length account. Together with the matched-vs.-mismatched control, this suggests that imported procedural content matters. The control does not establish equivalence across prose, JSON, or Markdown styles, and the selected routes are not used to estimate the prevalence of positive or negative transfer.

\subsection{Cost Profile: Usually Cost-Neutral, Sometimes Wasteful}

\begin{figure}[t]
\centering
\includegraphics[width=0.49\columnwidth]{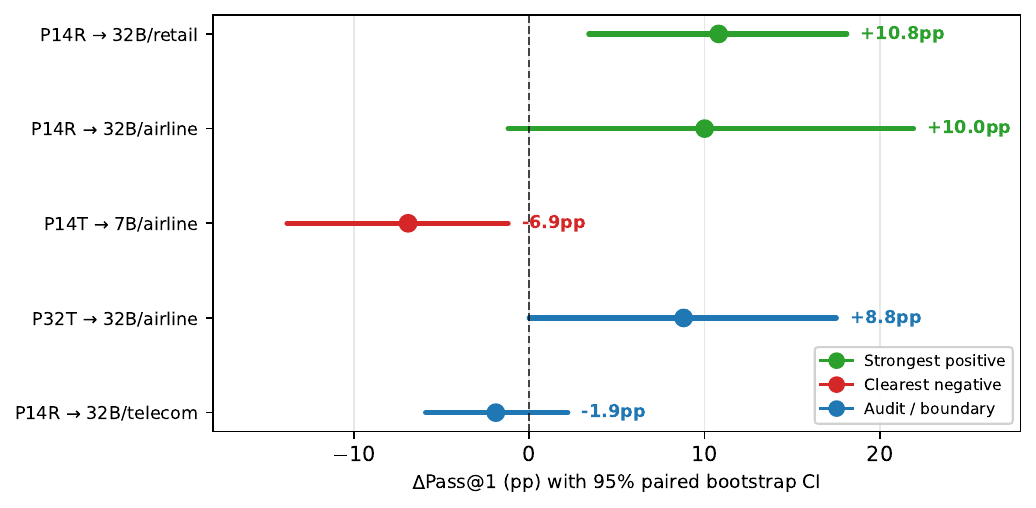}%
\hfill
\includegraphics[width=0.49\columnwidth]{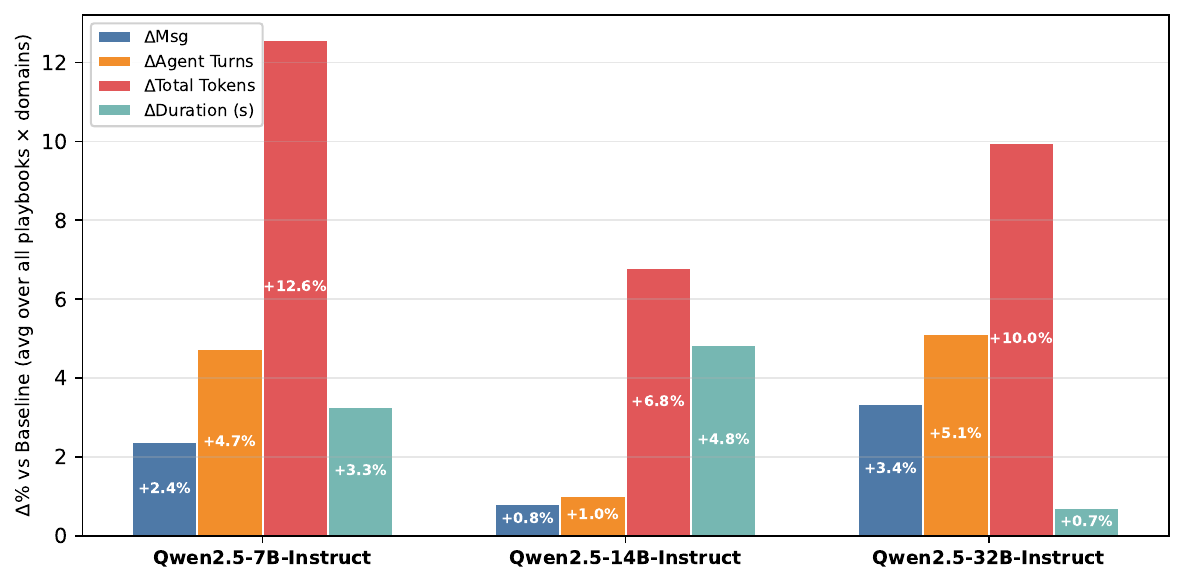}
\caption{TAU2-Bench follow-up views. Left: representative $\Delta$Pass$^1$ routes with paired-bootstrap 95\% CIs. Right: average cost overhead vs.\ baseline.}
\label{fig:tau2_followups}
\end{figure}

A complementary cost audit adds an important deployment-side clarification. Unlike ALFWorld, the main TAU2-Bench story is not large interaction-budget savings. When transfer helps, it is usually \emph{cost-neutral}: message counts and agent-turn counts change only slightly, and most token overhead comes from fixed prompt injection (Figure~\ref{fig:tau2_followups}, right).

Same-domain positive configurations often have flat or slightly shorter trajectories than baseline. Cross-family telecom gains usually come with higher token overhead, driven more by extra agent turns than by the fixed playbook text itself. Negative configurations show the opposite pattern: performance drops while cost rises. TAU2-Bench therefore contributes what ALFWorld cannot: transfer in service-agent settings is heterogeneous rather than smooth, and deployment requires validation and routing rather than blind reuse.

\subsection{Frozen Transfer versus Target-Side Redistillation}

We reorganize an existing pre-submission L1 source-model analysis around the deployment choice requested during review. For the same Qwen2.5-14B target, Table~\ref{tab:tau2_redistill_main} compares no playbook, a target-derived same-domain artifact, and same-domain artifacts transferred from 7B or 32B sources. No strategy dominates across all three domains, and interaction costs remain similar within domain. These are exploratory secondary estimands rather than additions to the 135-route confirmatory family; direct paired intervals are reported in Appendix~\ref{sec:app_tau2_redistill}.

\begin{table}[t]
\centering
\scriptsize
\setlength{\tabcolsep}{3.2pt}
\begin{tabular}{@{}lrrrr@{}}
\toprule
Domain & Base & Target & 7B source & 32B source \\
\midrule
Airline & 21.25 & 33.13 & 37.50 & 23.13 \\
Retail & 54.58 & 62.92 & 56.25 & 63.33 \\
Telecom & 56.25 & 58.44 & 59.38 & 62.19 \\
\bottomrule
\end{tabular}
\caption{Task-macro Pass$^1$ (\%) for a Qwen2.5-14B target with no playbook, a target-derived same-domain L1, or frozen same-domain L1 transfer. Results are exploratory.}
\label{tab:tau2_redistill_main}
\end{table}

Frozen transfer is therefore a viable cold-start candidate when target-side failures are unavailable or costly to collect, not a preferred default. Once target trajectories exist, transferred and target-derived artifacts should be compared on target-side accuracy, termination, protocol compatibility, and cost.

\section{Runtime Shift Can Destabilize Operational Behavior (XBench-DeepSearch)}

XBench-DeepSearch is our smallest benchmark and evaluates one primary frozen artifact across runtime conditions. It is an exploratory stress test of whether that artifact remains behaviorally bounded after runtime shift, not evidence for a general law of long-context deployment.

\subsection{Accuracy Is Mixed, but Operational Behavior Destabilizes}

\begin{figure}[t]
\centering
\includegraphics[width=\columnwidth]{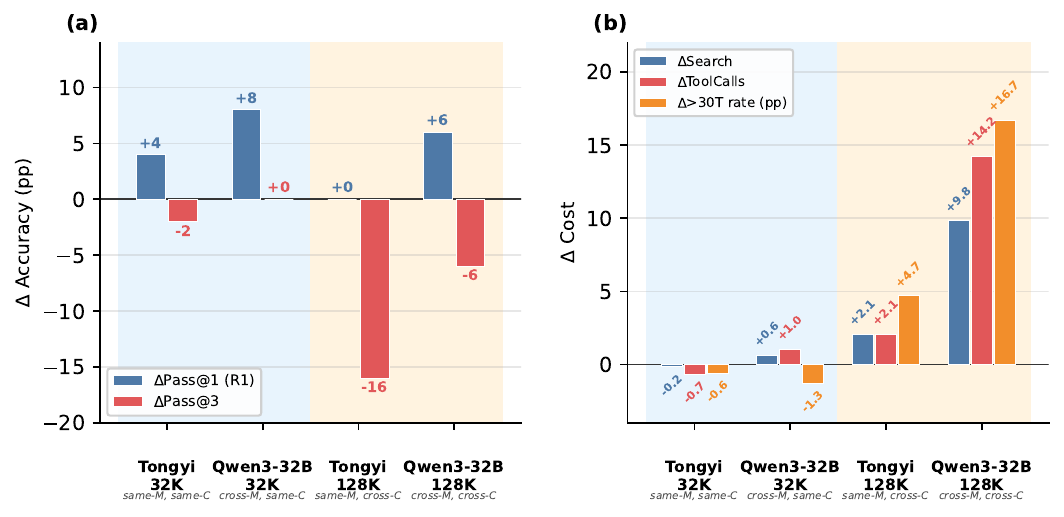}
\caption{Portability profile of the transferred XBench playbook across four deployment conditions. Left: $\Delta$Pass@1 and $\Delta$Pass@3. Right: $\Delta$Search, $\Delta$Tool, and $\Delta$($>$30T). M = model; C = context.}
\label{fig:xbench_portability}
\end{figure}

Figure~\ref{fig:xbench_portability} evaluates one frozen XBench playbook, distilled from Tongyi-DeepResearch-30B-A3B 32K source-side trajectories, across four deployment conditions: Tongyi-DR 32K, Qwen3-32B 32K, Tongyi-DR 128K, and Qwen3-32B 128K. Under same-context transfer, Pass@1 rises by $+4$ to $+8$ pp and Pass@3 is flat or only slightly lower. Under cross-context transfer, Pass@1 is mixed ($0$ to $+6$ pp) but Pass@3 falls by $6$ to $16$ pp. We therefore treat XBench as a runtime-change stress test rather than a benchmark-level verdict on portability.

The key value of XBench is cost-side behavior under runtime change. Under Tongyi-DR 32K, the transferred playbook is roughly cost-neutral. Under Tongyi-DR 128K, Pass@1 is flat but Pass@3 drops by 16 pp as search and tool calls rise. Under Qwen3-32B 128K, the same artifact becomes much less bounded: search calls rise from 5.9 to 15.7, total tool calls from 10.4 to 24.7, and overlong-trajectory rate ($>$30T) from 8\% to 25\%.

A 150-pair behavioral audit localizes the Qwen3-32B 128K shift: trajectories with an identical repeated query rise from 17.3\% to 39.3\%, those with at least three identical searches rise from 8\% to 24\%, and submissions within ten tool calls fall from 78\% to 30\%. Question-cluster intervals exclude zero, whereas corresponding 32K changes are smaller and cross zero. This supports re-querying and delayed stopping as a proximal behavioral pathway, not an internal causal mechanism (Appendix~\ref{sec:app_xbench_mech}).

Target-side redistillation does not automatically remove this risk. It leaves Qwen3-32B 128K accuracy statistically unresolved while raising average total tool calls from 10.4 to 36.1 ($+247\%$), with about 24\% of trajectories reaching a budget (Appendix~\ref{sec:app_xbench_redistill}).

Compatibility screening remains inconclusive. In a prospective Gemini-2.5-Pro pilot with a frozen $+5$ mean-tool-call alert, smoke and held-out increases are $+0.26$ and $+4.86$; both narrowly agree on no alert, but the 347.1\% held-out inflation demonstrates neither calibration nor a validated selector (Appendix~\ref{sec:app_xbench_precheck}).

\section{Deployment Lessons}
\label{sec:deployment_lessons}

Across benchmarks, transferred and target-derived playbooks are candidate artifacts rather than defaults. Deployment should freeze a target slice and risk criteria, compare accuracy, termination, protocol compatibility, and cost, then retain, repair, redistill, or reject the artifact. Cost is benchmark-dependent: ALFWorld can shorten failed rollouts, TAU2 gains are often cost-neutral, and XBench can turn inflation into the failure mode.

\paragraph{Checklist.}
\begin{itemize}
\setlength{\itemsep}{0pt}
\setlength{\topsep}{1pt}
\setlength{\parsep}{0pt}
\setlength{\partopsep}{0pt}
\item \textbf{Validate transfer and redistillation on target-side slices}, not source performance alone; prefer domain-matched candidates.
\item \textbf{Audit accuracy, termination, protocol compatibility, and cost}, especially for strong targets or runtime changes.
\item \textbf{Freeze decision criteria and keep a rollback path}; our pilot does not provide a universal threshold or selector.
\end{itemize}

Prompt-side playbooks are deployment artifacts, not universally reusable knowledge modules.

\section{Conclusion}
\label{sec:conclusion}

Prompt-side playbooks transfer only conditionally. They help in controlled settings and beat fixed demonstrations in one comparison; TAU2 shows a modest matched-domain advantage but little corrected route-level evidence; and one XBench pairing exhibits re-querying, delayed stopping, and cost inflation. Redistillation is not automatically safer, but frozen transfer is not a preferred default. The contribution is diagnostic: both require target-side validation before rollout.

\newpage
\section*{Limitations}
\label{sec:limitations}

This is an empirical Industry Track deployment study, not a new learning algorithm, validated compatibility selector, or internal-mechanism study. TAU2-Bench has limited per-cell power. Global Holm correction over its 135 Pass$^1$ routes retains only one positive route; family-specific Holm additionally retains two negative cross-family routes, while global BH retains all three only as an FDR sensitivity analysis. Other route estimates, Pass$^{2\text{--}4}$, and costs are descriptive, and failure to survive correction is not evidence of zero effect. The separately defined matched-versus-mismatched contrast supports only a modest average advantage. The three-route placebo weakens a pure length explanation but is not a benchmark-wide causal intervention or a full prose/JSON/Markdown format ablation.

The few-shot result compares one fixed five-example prompt with one distilled artifact in a compatible ALFWorld setting; it does not establish cross-paradigm superiority. We do not compare against human-authored playbooks or deployment-time dynamic retrieval systems. Our target-derived analyses are secondary comparisons: the TAU2 L1 slice was originally collected for source-model analysis, and the XBench comparison is small. Neither transfer nor target-side redistillation is uniformly preferable.

XBench-DeepSearch contains 100 public problems and the controlled runtime-shift result centers on one primary artifact--runtime pairing rather than independent artifact replications. The observed repeated querying and delayed stopping provide a proximal behavioral pathway, not evidence about stopping-probability calibration, attention, position effects, or other internal context-length-OOD mechanisms. The prospective cost-screening pilot covers one new model--artifact pairing and one deployment-dependent threshold; it provides no calibrated remedy, accuracy predictor, or general selector.

Finally, the evaluated domains emphasize embodied, service, and search agents; software engineering, code generation, and data analysis remain untested. ALFWorld uses Claude Code--Claude Opus 4.6 and Codex--GPT-5.4 implementations on a shared source pool, reducing dependence on one implementation but not establishing generalization beyond the common natural-language distill--validate--transfer workflow.

\section*{Acknowledgments}

We thank the anonymous reviewers for their detailed and constructive feedback.

\bibliography{custom}

\appendix

\section{Appendix Overview}

These appendices provide protocol and artifact details, benchmark-specific statistical audits, control analyses, and illustrative behavioral cases. Appendix~\ref{sec:app_protocol} summarizes the shared workflow, benchmark settings, metrics, and artifact inventories. Appendix~\ref{sec:app_alfworld} reports the ALFWorld lift tables, temperature robustness, and qualitative cases. Appendix~\ref{sec:app_tau2} reports TAU2 multiplicity correction, full descriptive grids, domain controls, target-derived comparisons, and placebo analyses. Appendix~\ref{sec:app_xbench} reports descriptive portability, target-derived and cost-screening comparisons, the runtime-shift behavioral audit, and an illustrative failure case.

\section{Experimental Protocol and Artifact Details}
\label{sec:app_protocol}

This section consolidates the cross-benchmark details that would otherwise be repeated in each audit: the shared distillation workflow, the benchmark/target setup, the metric conventions, and the concrete artifacts transferred in the main paper.

\subsection{Distill--Validate--Transfer Workflow}

Across all three benchmarks, the transferred object is a frozen prompt-side playbook distilled from repeated source-side trajectories. We first collect all-fail and mixed-outcome bundles from a source condition, ask a stronger closed-source distiller to abstract recurring procedures and recovery heuristics, validate the resulting playbook on the source side, and then freeze it before transfer. The transfer question is therefore always about portability of a fixed prompt artifact rather than target-side prompt retuning.

\subsection{Benchmarks, Targets, and Metrics}

ALFWorld serves as the controlled existence test, TAU2-Bench as the main heterogeneous service-agent benchmark, and XBench-DeepSearch as the runtime-shift stress test. On ALFWorld, we report normalized success rate with environment-broken games removed from the denominator. On TAU2-Bench, we use corrected official-style Pass$^k$ with Pass$^1$ as the headline metric and Pass$^{2\text{--}4}$ as reliability complements; route-level uncertainty is reported with paired bootstrap CI$_{95}$. On XBench-DeepSearch, we report Pass@1, Pass@3, and deployment-side cost metrics, because the main failure mode under runtime shift is operational rather than purely top-line.

\subsection{Artifact Inventory and Example Playbook}
\label{app:artifact_inventory}

To make the transferred objects less black-box, we inventory the main-result
artifacts by distiller model, harness, and surface form, and then provide one
complete example playbook plus one short cross-stack excerpt. Word counts are
whitespace-tokenized after stripping markdown headers. Rule counts measure
top-level numbered or headed directives only. Style is a qualitative
surface-form descriptor added for inspection and is not used in any
quantitative analysis.

\begin{table}[t]
\centering
\scriptsize
\setlength{\tabcolsep}{3pt}
\resizebox{\columnwidth}{!}{%
\begin{tabular}{@{}lrrrlll@{}}
\toprule
\textbf{Artifact} & \textbf{Words} & \textbf{Rules} & \textbf{Source} & \textbf{Distiller} & \textbf{Harness} & \textbf{Style} \\
\midrule
C7S  & 330 & 6  & Qwen2.5-7B-Instruct  & claude-opus-4.6 & Claude Code & procedural checklist \\
C32S & 478 & 7  & Qwen2.5-32B-Instruct & claude-opus-4.6 & Claude Code & procedural checklist \\
X7D  & 698 & 10 & Qwen2.5-7B-Instruct  & gpt-5.4         & Codex       & search-heavy guide \\
X7C  & 292 & 6  & Qwen2.5-7B-Instruct  & gpt-5.4         & Codex       & dense checklist \\
\bottomrule
\end{tabular}}
\caption{ALFWorld artifacts used in the main results. Naming follows
Section~\ref{sec:artifact_naming}; the source column lists the full
source model rather than only its size shorthand.}
\label{tab:app_artifact_alf}
\end{table}

\begin{table*}[t]
\centering
\small
\setlength{\tabcolsep}{4pt}
\begin{tabular}{@{}lrrrllll@{}}
\toprule
\textbf{Artifact} & \textbf{Words} & \textbf{Rules} & \textbf{Source} & \textbf{Domain} & \textbf{Distiller} & \textbf{Harness} & \textbf{Style} \\
\midrule
P7A  & 549 & 7  & Qwen2.5-7B-Instruct  & airline & claude-opus-4.6 & Claude Code & procedural checklist \\
P7R  & 502 & 10 & Qwen2.5-7B-Instruct  & retail  & claude-opus-4.6 & Claude Code & procedural checklist \\
P7T  & 476 & 8  & Qwen2.5-7B-Instruct  & telecom & claude-opus-4.6 & Claude Code & diagnostic guide \\
P14A & 319 & 7  & Qwen2.5-14B-Instruct & airline & claude-opus-4.6 & Claude Code & procedural checklist \\
P14R & 324 & 8  & Qwen2.5-14B-Instruct & retail  & claude-opus-4.6 & Claude Code & procedural checklist \\
P14T & 357 & 8  & Qwen2.5-14B-Instruct & telecom & claude-opus-4.6 & Claude Code & diagnostic guide \\
P32A & 305 & 7  & Qwen2.5-32B-Instruct & airline & claude-opus-4.6 & Claude Code & procedural checklist \\
P32R & 338 & 8  & Qwen2.5-32B-Instruct & retail  & claude-opus-4.6 & Claude Code & procedural checklist \\
P32T & 339 & 8  & Qwen2.5-32B-Instruct & telecom & claude-opus-4.6 & Claude Code & diagnostic guide \\
\bottomrule
\end{tabular}
\caption{TAU2-Bench playbooks used in the transfer study. These playbooks
are written in a domain-general surface form but are derived from
source-domain failure patterns.}
\label{tab:app_artifact_tau2}
\end{table*}

\begin{table}[t]
\centering
\scriptsize
\setlength{\tabcolsep}{3pt}
\resizebox{\columnwidth}{!}{%
\begin{tabular}{@{}lrrrlll@{}}
\toprule
\textbf{Artifact} & \textbf{Words} & \textbf{Rules} & \textbf{Source} & \textbf{Distiller} & \textbf{Harness} & \textbf{Style} \\
\midrule
XBench-32K & 478 & 8 & Tongyi-DeepResearch-30B-A3B 32K & claude-opus-4.6 & Claude Code & search protocol \\
\bottomrule
\end{tabular}}
\caption{XBench-DeepSearch artifact used in the main text. The transferred
XBench playbook was distilled from Tongyi-DeepResearch-30B-A3B 32K
source-side trajectories and includes explicit budget and stopping guidance.}
\label{tab:app_artifact_xbench}
\end{table}

\subsection{Example Playbook: P14R}
\label{sec:app_artifact_p14r}

Below we show the complete content of artifact P14R, the
Qwen2.5-14B-Instruct retail playbook used in the TAU2-Bench main
results. Rule content is unchanged; only whitespace and rule labels are
normalized for presentation. The verbatim content below preserves the
artifact's original internal shorthand.

\begin{Verbatim}[fontsize=\scriptsize]
# Customer-Service Agent Playbook -- P14R

These rules were abstracted from retail failure patterns. Apply every
rule below before any tool call that mutates state.

## R1 -- Read thoroughly before writing

In 14B retail, successes used read tools 2.53x/trial vs failures at
1.14x. Before every write:
1. Read the user's profile/details.
2. Read the target entity (order, reservation, account).
3. Read any referenced secondary entities.

If you have not called at least 2 read tools, you probably have not
read enough.

## R2 -- Authenticate before everything

Positively identify the customer via lookup tools before any action.
Never proceed without confirmed identity.

## R3 -- Lifecycle gates: check entity status

Each write requires the target entity to be in a specific state.
Check status from the most recent read-tool output. If status does not
match the write's precondition, refuse.

## R4 -- Verify all entity references in write arguments

Before a write that references other entities:
1. Verify each referenced entity exists via its own read tool.
2. Confirm IDs come from tool output, not user claims or memory.
3. Check new values differ from old values where required.

## R5 -- Confirm before write

Before every mutating tool call, state the action, entity IDs, and
effect. Ask for explicit confirmation.

## R6 -- Turn atomicity

One tool call per turn. Never combine speech and tool call.

## R7 -- Do not give up: retry with corrected arguments

If a tool call fails, do not transfer to human. Re-read the entity,
correct the arguments, and try again. Only transfer when the user
explicitly requests it.

## R8 -- Multi-intent: serialize and re-read

Multiple changes should be handled one at a time. After each write,
re-read the entity before the next operation.
\end{Verbatim}

\subsection{Cross-Stack Excerpt: X7D}
\label{sec:app_artifact_x7d}

For comparison, we show a short excerpt from X7D, a GPT-5.4-distilled
ALFWorld artifact produced via Codex from the same Qwen2.5-7B-Instruct
source pool as C7S. The purpose is only to make visible that artifacts
produced by different distiller/harness stacks can differ in length,
granularity, and presentation style.

\begin{Verbatim}[fontsize=\scriptsize]
1. Plan around the object, not the destination.
Do not go to the target receptacle first unless you are already
holding the target object.
Default pattern:
- pick_place: find object -> take object
  -> go to destination -> move object
- clean_place: find object -> take object
  -> go to sink/basin -> clean object
  -> go to destination -> move object
- heat_place: find object -> take object
  -> go to microwave -> heat object
  -> go to destination -> move object
- cool_place: find object -> take object
  -> go to fridge -> cool object
  -> go to destination -> move object

3. Search open surfaces before deep container chains.
The failed runs overused drawers, shelves, and cabinets in long loops.
For small moveable objects, first search exposed surfaces:
- desk, sidetable, coffeetable,
  diningtable, bed, armchair,
  countertop
Only start opening drawers/cabinets after one sweep of the obvious
exposed places.

7. Use exact target matching.
Do not substitute near-miss objects:
- mug is not cup
- pen is not pencil
- soapbar is not soapbottle
- cellphone is not laptop
Before taking an object, verify the noun exactly matches
the task target.

10. Lightweight room priors for this split.
Use these as first guesses, not as guarantees:
- examine/small office objects: desk, sidetable,
  coffeetable, bed, armchair
- bathroom objects: toilet, countertop, cabinet,
  handtowelholder,
  bathtubbasin
- kitchen objects: countertop, cabinet,
  diningtable, fridge
\end{Verbatim}

\section{ALFWorld Statistical Audit}
\label{sec:app_alfworld}

This section mirrors the ALFWorld story in the main text: we first report the full lift tables, then the temperature-robustness audit, and finally a small set of illustrative cases that make the main qualitative correction patterns concrete.

We report the latest 400-run audit: for Qwen2.5-7B-Instruct, Qwen2.5-32B-Instruct, Qwen3-32B, and Llama3.1-70B-Inst we run 5 seeds at both $t{=}0$ and $t{=}0.3$. The four ALFWorld artifacts are C7S, C32S, X7D, and X7C, following the naming scheme in Section~\ref{sec:artifact_naming}. Stars denote paired t-test significance against baseline or across temperatures: * $p{<}0.05$, ** $p{<}0.01$, *** $p{<}0.001$.

\subsection{Full Lift Tables}

\begin{table*}[t]
\centering
\small
\resizebox{\textwidth}{!}{%
\begin{tabular}{@{}llrrrrr@{}}
\toprule
\textbf{Target} & \textbf{Split} & \textbf{Base} & \textbf{C7S} & \textbf{C32S} & \textbf{X7D} & \textbf{X7C} \\
\midrule
\multicolumn{7}{@{}l}{\textit{$t{=}0$ (5-run mean)}} \\
\midrule
Qwen2.5-7B-Instruct  & seen   & 35.0 & +4.4*** & +6.3**  & +6.9**  & $-$0.1 \\
Qwen2.5-7B-Instruct  & unseen & 34.5 & +0.9    & +3.1    & +9.3*** & +2.3 \\
Qwen2.5-32B-Instruct & seen   & 67.9 & +4.9**  & +6.4*** & +9.9*** & +7.1*** \\
Qwen2.5-32B-Instruct & unseen & 62.7 & +11.6*** & +10.1*** & +12.2*** & +11.8*** \\
Qwen3-32B   & seen   & 43.0 & +11.2*** & +12.7** & +19.4*** & +22.4*** \\
Qwen3-32B   & unseen & 45.5 & +12.6*** & +13.5*** & +23.6*** & +16.2*** \\
Llama3.1-70B-Inst  & seen   & 40.4 & +15.1** & +15.1** & +22.4*** & +22.8*** \\
Llama3.1-70B-Inst  & unseen & 42.0 & +17.2** & +13.6** & +19.5*** & +15.3*** \\
\midrule
\multicolumn{7}{@{}l}{\textit{$t{=}0.3$ (5-run mean)}} \\
\midrule
Qwen2.5-7B-Instruct  & seen   & 36.2 & +4.9*   & +5.1**  & $-$16.0*** & $-$22.6** \\
Qwen2.5-7B-Instruct  & unseen & 27.6 & +0.5    & $-$1.8  & $-$7.3    & +0.3 \\
Qwen2.5-32B-Instruct & seen   & 61.3 & +5.6**  & +6.7    & +16.3**   & +0.9 \\
Qwen2.5-32B-Instruct & unseen & 57.0 & +7.6**  & +10.9*** & +13.5**  & $-$3.8* \\
Qwen3-32B   & seen   & 49.6 & +12.0*** & +16.9*** & +23.0*** & +22.4*** \\
Qwen3-32B   & unseen & 53.4 & +9.5*** & +10.8** & +21.9*** & +18.0*** \\
Llama3.1-70B-Inst  & seen   & 45.4 & +20.1*** & +23.3*** & +25.0*** & +23.4*** \\
Llama3.1-70B-Inst  & unseen & 47.2 & +23.4*** & +19.3*** & +19.3*** & +15.0*** \\
\bottomrule
\end{tabular}
}
\caption{ALFWorld playbook lift over baseline (pp). Base = baseline success rate; all other entries are baseline-to-playbook lifts.}
\label{tab:app_alf_lifts}
\end{table*}

\subsection{Temperature-Robustness Audit}

\begin{table*}[t]
\centering
\small
\resizebox{\textwidth}{!}{%
\begin{tabular}{@{}llrrrrr@{}}
\toprule
\textbf{Target} & \textbf{Split} & \textbf{Baseline} & \textbf{C7S} & \textbf{C32S} & \textbf{X7D} & \textbf{X7C} \\
\midrule
Qwen2.5-7B-Instruct  & seen   & +1.2   & +1.7   & $-$0.0   & $-$21.7*** & $-$21.3** \\
Qwen2.5-7B-Instruct  & unseen & $-$6.9 & $-$7.3 & $-$11.8  & $-$23.5*** & $-$8.9 \\
Qwen2.5-32B-Instruct & seen   & $-$6.6 & $-$5.9* & $-$6.3** & $-$0.2     & $-$12.8* \\
Qwen2.5-32B-Instruct & unseen & $-$5.7** & $-$9.7** & $-$4.9* & $-$4.3   & $-$21.2*** \\
Qwen3-32B   & seen   & +6.6* & +7.3* & +10.9* & +10.2** & +6.6 \\
Qwen3-32B   & unseen & +7.8* & +4.7 & +5.1 & +6.1* & +9.6* \\
Llama3.1-70B-Inst  & seen   & +5.1*  & +10.1** & +13.3*  & +7.7*     & +5.7 \\
Llama3.1-70B-Inst  & unseen & +5.1** & +11.4*  & +10.8** & +5.0      & +4.9* \\
\bottomrule
\end{tabular}
}
\caption{Temperature effect in success rate, defined as SR@$t{=}0.3$ minus SR@$t{=}0$ (pp).}
\label{tab:app_alf_temp}
\end{table*}

The cost metrics track these robustness outcomes closely. The clearest failure case is Qwen2.5-7B-Instruct + X7D: on \texttt{seen}, hit-50 rises from 58.1\% to 79.8\% and mean trajectory length rises from 32.8 to 42.4 steps; on \texttt{unseen}, the same quantities move from 56.2\% to 79.7\% and from 32.7 to 43.0. By contrast, the robust case Qwen2.5-32B-Instruct + X7D stays nearly flat in cost while preserving gains (hit-50 22.1\%$\to$22.4\% and trajectory length 17.9$\to$18.1 on \texttt{seen}). Qwen3-32B and Llama3.1-70B-Inst show the opposite tendency: sampling lowers both hit-50 and trajectory length for baseline and playbook conditions, consistent with the positive temperature effect reported in Table~\ref{tab:app_alf_temp}.

\subsection{Illustrative ALFWorld Cases}

These cases are illustrative rather than statistical. We select them because they instantiate the dominant qualitative pattern behind the aggregate gains on ALFWorld: task-relevant search priors are corrected first, and task-specific action templates are corrected second. All three cases come from task types that also show positive aggregate transfer in the quantitative analysis (Qwen2.5-32B-Instruct, X7D, \texttt{valid\_seen}, $t{=}0$).

\begin{table*}[t]
\centering
\small
\resizebox{\textwidth}{!}{%
\begin{tabular}{@{}lllll@{}}
\toprule
\textbf{Task} & \textbf{Baseline} & \textbf{X7D} & \textbf{Dominant failure pattern} & \textbf{Correction mechanism} \\
\midrule
\texttt{clean\_place} & fail / 50 & success / 6  & target object seen, wrong action template selected & task-template correction \\
\texttt{heat\_place}  & fail / 50 & success / 10 & blind drawer/cabinet search; appliance never visited & search-priority correction \\
\texttt{cool\_place}  & fail / 50 & success / 12 & task-relevant locations ignored despite being observable & search-priority correction \\
\bottomrule
\end{tabular}
}
\caption{Illustrative ALFWorld cases from Qwen2.5-32B-Instruct with X7D on \texttt{valid\_seen} at $t{=}0$. These cases are chosen to expose the recurring qualitative pattern behind the aggregate gains, not as additional statistical evidence.}
\label{tab:app_alf_cases}
\end{table*}

\paragraph{clean\_place.} In the \texttt{soapbar $\rightarrow$ sinkbasin $\rightarrow$ bathtubbasin} task, the baseline agent spends most of its budget opening drawers, reaches the correct countertop only at step 26, observes the soapbar, and still picks up a cloth instead. It then executes a syntactically valid clean-and-place sequence on the wrong object. X7D goes directly to the countertop, takes the soapbar, cleans it at the sinkbasin, and places it in the bathtubbasin. This case highlights action-template correction rather than simple search acceleration.

\paragraph{heat\_place.} In the \texttt{egg $\rightarrow$ microwave $\rightarrow$ countertop} task, the baseline agent treats the episode as a generic container-search problem, opening long sequences of drawers and cabinets while never visiting the microwave or the actual egg location. X7D instead checks likely egg locations first, retrieves the egg, and completes the required heat-and-place sequence. The improvement comes from redirecting search toward task-relevant locations, not from removing search entirely.

\paragraph{cool\_place.} In the \texttt{pan $\rightarrow$ fridge $\rightarrow$ diningtable} task, the initial observation already exposes stoveburners, the fridge, and the diningtable, yet the baseline agent spends 34 of 50 actions on cabinets and never visits any of the task-relevant locations. X7D checks countertops, then stoveburners, retrieves the pan, cools it in the fridge, and places it on the diningtable. This case again shows that the playbook helps mainly by changing what the agent searches first.

Taken together, these cases support the mechanism suggested by the aggregate cost pattern: ALFWorld gains come primarily from rescuing episodes that would otherwise exhaust the 50-step ceiling, rather than from dramatically shortening already-successful trajectories.

\section{TAU2-Bench Statistical Audit}
\label{sec:app_tau2}

This section begins with the frozen eight-trial multiplicity audit, then gives the full descriptive grids, the aggregate domain control, target-derived comparisons, and illustrative content controls.

\subsection{Additional-Trial Audit}

\begin{table*}[t]
\centering
\small
\begin{tabular}{@{}lrcrll@{}}
\toprule
\textbf{Route} & \textbf{8-trial $\Delta$} & \textbf{8-trial CI} & \textbf{Latest $\Delta$} & \textbf{Latest CI} & \textbf{Reading} \\
\midrule
P32T $\to$ Qwen2.5-32B-Instruct/airline & +8.8 & [+0.0, +17.5] & +8.4 & [+1.6, +15.9] & positive CI \\
P14R $\to$ Qwen2.5-32B-Instruct/airline & +10.0 & [$-$1.2, +21.9] & +9.7 & [+0.9, +20.0] & positive CI \\
P7T $\to$ Qwen2.5-32B-Instruct/airline  & +8.1 & [$-$1.9, +18.8] & +8.8 & [+0.6, +18.4] & positive CI \\
P7A $\to$ Qwen2.5-14B-Instruct/retail  & +5.6 & [+0.0, +11.7] & +3.1 & [$-$0.4, +7.2] & overlaps 0 \\
P32A $\to$ Qwen2.5-14B-Instruct/retail & +3.5 & [$-$1.2, +8.2] & +3.0 & [$-$1.8, +8.0] & overlaps 0 \\
P14R $\to$ Qwen2.5-7B-Instruct/telecom  & +6.9 & [$-$0.3, +13.4] & +5.2 & [$-$1.1, +11.6] & overlaps 0 \\
\bottomrule
\end{tabular}
\caption{Descriptive additional-trial audit. Trials appended after the submitted eight-trial snapshot are a separate sensitivity analysis and are excluded from the 135-route multiplicity family.}
\label{tab:app_tau2_audit}
\end{table*}

\subsection{Multiplicity-Corrected Route Summary}

\begin{table}[t]
\centering
\scriptsize
\resizebox{\columnwidth}{!}{%
\begin{tabular}{@{}llrrrr@{}}
\toprule
\textbf{Route} & \textbf{$\Delta$} & \textbf{Fam. H} & \textbf{Fam. BH} & \textbf{Glob. H} & \textbf{Glob. BH} \\
\midrule
P14T$\to$Q2.5-7B/T & +9.69 & .0196 & .0196 & .0327 & .0316 \\
P32T$\to$Q3-8B/T & $-$9.38 & .0299 & .0190 & .0742 & .0316 \\
P7A$\to$Q3-32B/R & $-$9.48 & .0372 & .0190 & .0934 & .0316 \\
\bottomrule
\end{tabular}
}
\caption{Pass$^1$ routes retained by family-specific Holm (H) or global BH correction. Global Holm over all 135 routes retains only P14T$\to$Q2.5-7B/telecom. BH is reported as an FDR sensitivity analysis.}
\label{tab:app_tau2_supported}
\end{table}

The 81-route in-family Holm family retains only the first positive route. The 54-route cross-family family retains the latter two negative routes. Those negative routes do not survive the more conservative 135-route global Holm correction.

\subsection{Full In-Family and Cross-Family Transfer Tables}

The full numeric tables for both in-family and cross-family transfer are given below. Table~\ref{tab:app_tau2_l2_full} covers the 81 in-family configurations; Table~\ref{tab:app_tau2_cross_l2_full} covers the 54 cross-family configurations. For compactness, these full tables retain the original internal configuration IDs used during artifact generation; the main text and representative-case discussions use the shortened naming scheme from Section~\ref{sec:artifact_naming}.

\begin{table*}[t]
\centering
\tiny
\setlength{\tabcolsep}{1.2pt}
\resizebox{\textwidth}{!}{%
\begin{tabular}{@{}lll r r r l r r r l r r r l r r r l@{}}
\toprule
& & & \multicolumn{4}{c}{\textbf{Pass$^1$}} & \multicolumn{4}{c}{\textbf{Pass$^2$}} & \multicolumn{4}{c}{\textbf{Pass$^3$}} & \multicolumn{4}{c}{\textbf{Pass$^4$}} \\
\cmidrule(lr){4-7} \cmidrule(lr){8-11} \cmidrule(lr){12-15} \cmidrule(lr){16-19}
\textbf{Src} & \textbf{Tgt} & \textbf{Dom} & \textbf{BL} & \textbf{PB} & \textbf{$\Delta$} & \textbf{CI$_{95}$} & \textbf{BL} & \textbf{PB} & \textbf{$\Delta$} & \textbf{CI$_{95}$} & \textbf{BL} & \textbf{PB} & \textbf{$\Delta$} & \textbf{CI$_{95}$} & \textbf{BL} & \textbf{PB} & \textbf{$\Delta$} & \textbf{CI$_{95}$} \\
\midrule
7b\_a & Qwen2.5-7B-Instruct & airline & 15.6 & 16.2 & +0.6 & [$-$3.8, +5.0] & 8.4 & 6.3 & $-$2.1 & [$-$6.1, +0.4] & 4.8 & 2.9 & $-$2.0 & [$-$5.4, +0.0] & 2.6 & 1.3 & $-$1.3 & [$-$3.6, +0.0] \\
7b\_a & Qwen2.5-7B-Instruct & retail & 21.5 & 21.9 & +0.4 & [$-$4.8, +5.5] & 11.1 & 10.7 & $-$1.2 & [$-$6.7, +3.9] & 7.3 & 7.0 & $-$0.9 & [$-$5.5, +3.3] & 4.9 & 5.6 & +0.3 & [$-$3.2, +5.1] \\
7b\_a & Qwen2.5-7B-Instruct & telecom & 50.0 & 52.2 & +2.2 & [$-$3.1, +7.2] & 37.7 & 36.7 & $-$1.0 & [$-$6.6, +4.2] & 30.9 & 29.4 & $-$1.5 & [$-$7.4, +4.1] & 26.3 & 25.5 & $-$0.8 & [$-$6.2, +4.5] \\
7b\_r & Qwen2.5-7B-Instruct & airline & 15.6 & 13.8 & $-$1.9 & [$-$7.5, +3.8] & 8.4 & 3.8 & $-$4.6$\bigstar$ & [$-$9.6, $-$0.5] & 4.8 & 1.1 & $-$3.8$\bigstar$ & [$-$8.0, $-$0.4] & 2.6 & 0.2 & $-$2.4$\bigstar$ & [$-$5.3, $-$0.2] \\
7b\_r & Qwen2.5-7B-Instruct & retail & 21.5 & 22.1 & +0.0 & [$-$5.5, +5.5] & 11.1 & 11.6 & $-$0.2 & [$-$6.0, +5.1] & 7.3 & 8.2 & +0.6 & [$-$4.0, +5.6] & 4.9 & 6.9 & +2.0 & [$-$2.4, +7.8] \\
7b\_r & Qwen2.5-7B-Instruct & telecom & 50.0 & 55.9 & +5.9 & [$-$0.3, +12.2] & 37.7 & 38.8 & +1.2 & [$-$5.0, +7.1] & 30.9 & 30.9 & $-$0.0 & [$-$6.2, +6.3] & 26.3 & 26.2 & $-$0.1 & [$-$6.0, +6.5] \\
7b\_t & Qwen2.5-7B-Instruct & airline & 15.6 & 15.0 & $-$0.6 & [$-$6.2, +5.0] & 8.4 & 5.4 & $-$3.0 & [$-$8.2, +0.4] & 4.8 & 2.3 & $-$2.5 & [$-$6.4, +0.0] & 2.6 & 1.1 & $-$1.4 & [$-$4.1, +0.0] \\
7b\_t & Qwen2.5-7B-Instruct & retail & 21.5 & 23.1 & $-$0.4 & [$-$6.6, +6.2] & 11.1 & 11.1 & $-$1.2 & [$-$6.8, +4.5] & 7.3 & 7.1 & $-$0.7 & [$-$5.7, +4.3] & 4.9 & 5.3 & +0.0 & [$-$4.2, +5.2] \\
7b\_t & Qwen2.5-7B-Instruct & telecom & 50.0 & 56.2 & +6.2 & [$-$0.9, +12.8] & 37.7 & 37.3 & $-$0.4 & [$-$6.6, +5.5] & 30.9 & 28.5 & $-$2.4 & [$-$8.4, +3.2] & 26.3 & 24.0 & $-$2.3 & [$-$7.6, +2.8] \\
14b\_a & Qwen2.5-7B-Instruct & airline & 15.6 & 18.1 & +2.5 & [$-$3.1, +8.8] & 8.4 & 8.0 & $-$0.4 & [$-$5.4, +4.1] & 4.8 & 3.2 & $-$1.6 & [$-$5.4, +1.2] & 2.6 & 1.0 & $-$1.6 & [$-$4.3, +0.2] \\
14b\_a & Qwen2.5-7B-Instruct & retail & 21.5 & 17.4 & $-$4.3 & [$-$9.6, +0.4] & 11.1 & 7.5 & $-$3.9 & [$-$8.8, +0.3] & 7.3 & 4.0 & $-$3.8 & [$-$8.5, +0.1] & 4.9 & 2.0 & $-$3.3 & [$-$7.4, +0.1] \\
14b\_a & Qwen2.5-7B-Instruct & telecom & 50.0 & 51.9 & +1.9 & [$-$3.8, +8.1] & 37.7 & 36.1 & $-$1.6 & [$-$7.8, +4.6] & 30.9 & 28.5 & $-$2.4 & [$-$8.2, +3.5] & 26.3 & 24.1 & $-$2.2 & [$-$7.3, +3.0] \\
14b\_r & Qwen2.5-7B-Instruct & airline & 15.6 & 19.4 & +3.8 & [$-$2.5, +9.4] & 8.4 & 7.1 & $-$1.2 & [$-$6.1, +2.5] & 4.8 & 3.2 & $-$1.6 & [$-$5.5, +1.2] & 2.6 & 1.5 & $-$1.1 & [$-$3.4, +0.6] \\
14b\_r & Qwen2.5-7B-Instruct & retail & 21.5 & 20.5 & $-$1.1 & [$-$7.0, +4.8] & 11.1 & 12.7 & +0.9 & [$-$4.5, +6.7] & 7.3 & 8.6 & +1.0 & [$-$3.4, +5.4] & 4.9 & 6.3 & +1.4 & [$-$2.0, +5.9] \\
14b\_r & Qwen2.5-7B-Instruct & telecom & 50.0 & 56.9 & +6.9 & [$-$0.3, +13.4] & 37.7 & 40.4 & +2.8 & [$-$4.1, +9.5] & 30.9 & 33.2 & +2.3 & [$-$4.7, +10.0] & 26.3 & 29.3 & +3.0 & [$-$3.4, +10.6] \\
14b\_t & Qwen2.5-7B-Instruct & airline & 15.6 & 8.8 & $-$6.9$\bigstar$ & [$-$13.8, $-$1.2] & 8.4 & 2.9 & $-$5.5$\bigstar$ & [$-$11.1, $-$0.9] & 4.8 & 1.1 & $-$3.8$\bigstar$ & [$-$8.3, $-$0.5] & 2.6 & 0.4 & $-$2.2$\bigstar$ & [$-$5.0, $-$0.1] \\
14b\_t & Qwen2.5-7B-Instruct & retail & 21.5 & 19.3 & $-$1.8 & [$-$8.5, +4.0] & 11.1 & 10.3 & $-$2.1 & [$-$7.8, +2.6] & 7.3 & 6.4 & $-$1.7 & [$-$5.7, +1.4] & 4.9 & 4.1 & $-$1.2 & [$-$3.4, +0.5] \\
14b\_t & Qwen2.5-7B-Instruct & telecom & 50.0 & 59.7 & +9.7$\bigstar$ & [+4.7, +15.0] & 37.7 & 44.3 & +6.6$\bigstar$ & [+2.1, +11.9] & 30.9 & 37.1 & +6.2$\bigstar$ & [+0.8, +12.4] & 26.3 & 32.8 & +6.5$\bigstar$ & [+0.0, +13.8] \\
32b\_a & Qwen2.5-7B-Instruct & airline & 15.6 & 18.8 & +3.1 & [$-$3.8, +9.4] & 8.4 & 9.3 & +0.9 & [$-$5.4, +7.1] & 4.8 & 5.5 & +0.7 & [$-$4.9, +7.0] & 2.6 & 3.6 & +1.1 & [$-$3.2, +6.4] \\
32b\_a & Qwen2.5-7B-Instruct & retail & 21.5 & 21.2 & $-$0.4 & [$-$3.9, +3.6] & 11.1 & 10.1 & $-$1.5 & [$-$4.5, +1.5] & 7.3 & 6.4 & $-$1.0 & [$-$4.5, +2.5] & 4.9 & 5.0 & $-$0.0 & [$-$4.1, +4.4] \\
32b\_a & Qwen2.5-7B-Instruct & telecom & 50.0 & 50.3 & +0.3 & [$-$5.0, +5.3] & 37.7 & 33.8 & $-$3.8 & [$-$8.5, +0.5] & 30.9 & 26.3 & $-$4.6$\bigstar$ & [$-$8.7, $-$0.6] & 26.3 & 21.9 & $-$4.4$\bigstar$ & [$-$8.4, $-$1.0] \\
32b\_r & Qwen2.5-7B-Instruct & airline & 15.6 & 11.2 & $-$4.4 & [$-$10.6, +1.2] & 8.4 & 3.6 & $-$4.8$\bigstar$ & [$-$11.1, $-$0.2] & 4.8 & 1.2 & $-$3.7$\bigstar$ & [$-$7.8, $-$0.3] & 2.6 & 0.4 & $-$2.2$\bigstar$ & [$-$4.8, $-$0.1] \\
32b\_r & Qwen2.5-7B-Instruct & retail & 21.5 & 23.7 & +3.5 & [$-$3.5, +10.4] & 11.1 & 12.6 & +1.1 & [$-$5.7, +7.9] & 7.3 & 8.0 & +0.9 & [$-$5.4, +8.5] & 4.9 & 7.1 & +2.2 & [$-$4.6, +10.9] \\
32b\_r & Qwen2.5-7B-Instruct & telecom & 50.0 & 58.4 & +8.4$\bigstar$ & [+2.5, +14.1] & 37.7 & 42.8 & +5.1 & [$-$0.6, +10.9] & 30.9 & 34.7 & +3.8 & [$-$1.6, +9.5] & 26.3 & 29.8 & +3.5 & [$-$2.4, +9.1] \\
32b\_t & Qwen2.5-7B-Instruct & airline & 15.6 & 14.4 & $-$1.2 & [$-$5.0, +2.5] & 8.4 & 7.3 & $-$1.1 & [$-$4.8, +2.5] & 4.8 & 4.5 & $-$0.4 & [$-$3.6, +3.5] & 2.6 & 2.9 & +0.4 & [$-$2.4, +4.0] \\
32b\_t & Qwen2.5-7B-Instruct & retail & 21.5 & 20.2 & $-$2.7 & [$-$8.3, +2.7] & 11.1 & 9.5 & $-$2.0 & [$-$6.1, +1.8] & 7.3 & 5.9 & $-$1.2 & [$-$5.2, +2.6] & 4.9 & 4.7 & $-$0.4 & [$-$4.4, +4.5] \\
32b\_t & Qwen2.5-7B-Instruct & telecom & 50.0 & 47.8 & $-$2.2 & [$-$7.8, +3.1] & 37.7 & 32.8 & $-$4.9 & [$-$10.8, +0.4] & 30.9 & 26.5 & $-$4.4 & [$-$10.4, +1.1] & 26.3 & 22.7 & $-$3.6 & [$-$9.1, +2.6] \\
\midrule
7b\_a & Qwen2.5-14B-Instruct & airline & 21.2 & 26.2 & +5.0 & [$-$1.9, +11.2] & 11.4 & 9.8 & $-$1.6 & [$-$10.0, +4.6] & 7.6 & 4.0 & $-$3.6 & [$-$13.1, +3.3] & 5.9 & 1.6 & $-$4.2 & [$-$14.5, +2.1] \\
7b\_a & Qwen2.5-14B-Instruct & retail & 48.2 & 58.5 & +5.6 & [+0.0, +11.7] & 35.6 & 42.7 & +7.1$\bigstar$ & [+0.2, +14.6] & 26.0 & 33.1 & +7.1 & [$-$0.6, +15.6] & 19.5 & 25.6 & +6.1 & [$-$1.8, +15.0] \\
7b\_a & Qwen2.5-14B-Instruct & telecom & 56.2 & 55.0 & $-$1.2 & [$-$6.6, +3.8] & 44.0 & 44.8 & +0.8 & [$-$5.0, +7.1] & 38.1 & 39.8 & +1.7 & [$-$5.8, +9.4] & 34.2 & 36.1 & +1.9 & [$-$7.1, +10.6] \\
7b\_r & Qwen2.5-14B-Instruct & airline & 21.2 & 24.4 & +3.1 & [$-$3.1, +9.4] & 11.4 & 11.1 & $-$0.4 & [$-$6.1, +7.0] & 7.6 & 7.1 & $-$0.4 & [$-$6.9, +7.4] & 5.9 & 5.1 & $-$0.7 & [$-$7.9, +6.6] \\
7b\_r & Qwen2.5-14B-Instruct & retail & 48.2 & 51.5 & +1.9 & [$-$3.4, +7.2] & 35.6 & 39.4 & +3.8 & [$-$2.6, +10.5] & 26.0 & 30.5 & +5.6 & [$-$1.7, +13.4] & 19.5 & 25.8 & +6.3 & [$-$1.0, +15.3] \\
7b\_r & Qwen2.5-14B-Instruct & telecom & 56.2 & 57.2 & +0.9 & [$-$5.0, +6.9] & 44.0 & 43.8 & $-$0.3 & [$-$7.9, +6.7] & 38.1 & 36.5 & $-$1.6 & [$-$10.4, +6.7] & 34.2 & 31.6 & $-$2.6 & [$-$12.5, +6.3] \\
7b\_t & Qwen2.5-14B-Instruct & airline & 21.2 & 21.9 & +0.6 & [$-$5.0, +6.2] & 11.4 & 10.2 & $-$1.2 & [$-$8.9, +6.2] & 7.6 & 6.2 & $-$1.3 & [$-$10.4, +7.4] & 5.9 & 4.0 & $-$1.9 & [$-$12.1, +6.9] \\
7b\_t & Qwen2.5-14B-Instruct & retail & 48.2 & 55.0 & +0.4 & [$-$5.8, +7.1] & 35.6 & 37.1 & +0.2 & [$-$5.2, +6.8] & 26.0 & 25.4 & $-$0.6 & [$-$6.1, +4.7] & 19.5 & 17.7 & $-$1.7 & [$-$7.8, +3.9] \\
7b\_t & Qwen2.5-14B-Instruct & telecom & 56.2 & 55.0 & $-$1.2 & [$-$5.9, +3.8] & 44.0 & 42.4 & $-$1.6 & [$-$8.0, +4.6] & 38.1 & 35.2 & $-$2.9 & [$-$10.4, +4.0] & 34.2 & 30.1 & $-$4.1 & [$-$12.8, +3.2] \\
14b\_a & Qwen2.5-14B-Instruct & airline & 21.2 & 25.6 & +4.4 & [$-$6.9, +15.0] & 11.4 & 9.6 & $-$1.8 & [$-$13.8, +7.3] & 7.6 & 3.8 & $-$3.8 & [$-$15.5, +4.0] & 5.9 & 1.4 & $-$4.4 & [$-$15.3, +2.1] \\
14b\_a & Qwen2.5-14B-Instruct & retail & 48.2 & 58.1 & +5.2 & [$-$2.0, +12.5] & 35.6 & 43.3 & +6.5 & [$-$1.1, +14.5] & 26.0 & 32.8 & +6.8 & [$-$0.1, +13.3] & 19.5 & 25.8 & +6.4$\bigstar$ & [+0.9, +12.0] \\
14b\_a & Qwen2.5-14B-Instruct & telecom & 56.2 & 57.8 & +1.6 & [$-$4.4, +7.2] & 44.0 & 44.7 & +0.7 & [$-$6.8, +8.5] & 38.1 & 38.5 & +0.4 & [$-$9.3, +9.5] & 34.2 & 34.5 & +0.3 & [$-$10.4, +11.4] \\
14b\_r & Qwen2.5-14B-Instruct & airline & 21.2 & 23.8 & +2.5 & [$-$7.5, +11.9] & 11.4 & 9.8 & $-$1.6 & [$-$13.9, +7.0] & 7.6 & 4.3 & $-$3.3 & [$-$15.2, +4.6] & 5.9 & 1.9 & $-$4.0 & [$-$15.2, +2.9] \\
14b\_r & Qwen2.5-14B-Instruct & retail & 48.2 & 59.3 & +6.5$\bigstar$ & [+2.0, +11.3] & 35.6 & 43.9 & +8.3$\bigstar$ & [+2.1, +14.3] & 26.0 & 35.3 & +9.3$\bigstar$ & [+1.9, +16.9] & 19.5 & 29.1 & +9.7$\bigstar$ & [+1.3, +18.8] \\
14b\_r & Qwen2.5-14B-Instruct & telecom & 56.2 & 56.6 & +0.3 & [$-$5.3, +5.9] & 44.0 & 45.4 & +1.4 & [$-$4.9, +7.8] & 38.1 & 39.5 & +1.4 & [$-$6.3, +9.5] & 34.2 & 35.4 & +1.1 & [$-$8.5, +10.6] \\
14b\_t & Qwen2.5-14B-Instruct & airline & 21.2 & 17.5 & $-$3.8 & [$-$13.8, +3.8] & 11.4 & 4.8 & $-$6.6 & [$-$17.9, +0.2] & 7.6 & 1.3 & $-$6.3$\bigstar$ & [$-$16.8, $-$0.1] & 5.9 & 0.4 & $-$5.5$\bigstar$ & [$-$15.7, $-$0.1] \\
14b\_t & Qwen2.5-14B-Instruct & retail & 48.2 & 50.4 & $-$0.8 & [$-$6.5, +5.6] & 35.6 & 35.6 & +0.0 & [$-$7.4, +8.0] & 26.0 & 26.1 & +1.0 & [$-$7.5, +9.2] & 19.5 & 21.1 & +1.7 & [$-$8.0, +11.0] \\
14b\_t & Qwen2.5-14B-Instruct & telecom & 56.2 & 57.8 & +1.6 & [$-$3.4, +6.9] & 44.0 & 44.6 & +0.6 & [$-$6.3, +7.5] & 38.1 & 37.6 & $-$0.5 & [$-$8.3, +7.9] & 34.2 & 32.9 & $-$1.3 & [$-$10.0, +7.4] \\
32b\_a & Qwen2.5-14B-Instruct & airline & 21.2 & 24.4 & +3.1 & [$-$1.9, +8.8] & 11.4 & 10.4 & $-$1.1 & [$-$6.8, +3.0] & 7.6 & 4.6 & $-$3.0 & [$-$10.1, +1.2] & 5.9 & 1.9 & $-$3.9 & [$-$12.4, +0.6] \\
32b\_a & Qwen2.5-14B-Instruct & retail & 48.2 & 54.7 & +3.5 & [$-$1.2, +8.2] & 35.6 & 41.0 & +5.4 & [$-$1.2, +12.5] & 26.0 & 32.0 & +6.0 & [$-$2.3, +15.4] & 19.5 & 25.3 & +5.9 & [$-$2.7, +16.5] \\
32b\_a & Qwen2.5-14B-Instruct & telecom & 56.2 & 58.8 & +2.5 & [$-$2.8, +7.8] & 44.0 & 48.8 & +4.7 & [$-$1.6, +11.3] & 38.1 & 44.4 & +6.3 & [$-$1.7, +14.5] & 34.2 & 41.3 & +7.1 & [$-$2.5, +16.9] \\
32b\_r & Qwen2.5-14B-Instruct & airline & 21.2 & 18.1 & $-$3.1 & [$-$11.9, +5.6] & 11.4 & 7.9 & $-$3.6 & [$-$13.0, +5.5] & 7.6 & 4.4 & $-$3.2 & [$-$14.1, +6.0] & 5.9 & 2.9 & $-$3.0 & [$-$14.6, +6.4] \\
32b\_r & Qwen2.5-14B-Instruct & retail & 48.2 & 50.7 & +2.7 & [$-$3.5, +9.0] & 35.6 & 37.7 & +4.6 & [$-$4.3, +14.2] & 26.0 & 32.3 & +6.3 & [$-$4.1, +18.1] & 19.5 & 26.8 & +7.3 & [$-$4.2, +20.6] \\
32b\_r & Qwen2.5-14B-Instruct & telecom & 56.2 & 58.8 & +2.5 & [$-$2.8, +7.8] & 44.0 & 47.6 & +3.6 & [$-$2.7, +10.1] & 38.1 & 41.1 & +3.0 & [$-$3.8, +9.7] & 34.2 & 36.5 & +2.2 & [$-$5.3, +9.8] \\
32b\_t & Qwen2.5-14B-Instruct & airline & 21.2 & 18.8 & $-$2.5 & [$-$11.2, +5.0] & 11.4 & 5.7 & $-$5.7 & [$-$15.9, +1.4] & 7.6 & 1.8 & $-$5.8 & [$-$17.5, +0.8] & 5.9 & 0.5 & $-$5.4 & [$-$15.9, +0.4] \\
32b\_t & Qwen2.5-14B-Instruct & retail & 48.2 & 54.8 & +2.0 & [$-$5.2, +8.9] & 35.6 & 38.5 & +3.0 & [$-$6.3, +12.1] & 26.0 & 29.2 & +3.3 & [$-$6.5, +13.1] & 19.5 & 22.6 & +3.1 & [$-$6.7, +12.5] \\
32b\_t & Qwen2.5-14B-Instruct & telecom & 56.2 & 56.9 & +0.6 & [$-$4.4, +5.3] & 44.0 & 45.1 & +1.1 & [$-$5.6, +7.5] & 38.1 & 39.7 & +1.7 & [$-$5.8, +9.5] & 34.2 & 36.0 & +1.7 & [$-$6.4, +10.9] \\
\midrule
7b\_a & Qwen2.5-32B-Instruct & airline & 26.9 & 31.9 & +5.0 & [$-$3.1, +13.8] & 12.7 & 20.0 & +7.3 & [$-$2.1, +18.4] & 6.5 & 13.5 & +7.0 & [$-$3.8, +18.5] & 3.5 & 9.8 & +6.3 & [$-$3.9, +18.4] \\
7b\_a & Qwen2.5-32B-Instruct & retail & 60.3 & 63.8 & +3.4 & [$-$3.4, +10.8] & 44.1 & 48.9 & +4.8 & [$-$4.8, +14.8] & 34.4 & 40.2 & +5.8 & [$-$6.7, +17.4] & 28.2 & 34.3 & +6.1 & [$-$7.9, +19.4] \\
7b\_a & Qwen2.5-32B-Instruct & telecom & 55.6 & 55.0 & $-$0.6 & [$-$5.6, +4.4] & 46.0 & 46.8 & +0.8 & [$-$5.1, +6.6] & 41.2 & 43.0 & +1.8 & [$-$5.3, +8.9] & 37.9 & 40.2 & +2.3 & [$-$6.8, +11.7] \\
7b\_r & Qwen2.5-32B-Instruct & airline & 26.9 & 29.4 & +2.5 & [$-$8.1, +13.8] & 12.7 & 17.5 & +4.8 & [$-$6.4, +16.8] & 6.5 & 11.5 & +5.0 & [$-$4.9, +14.7] & 3.5 & 7.6 & +4.1 & [$-$4.4, +12.4] \\
7b\_r & Qwen2.5-32B-Instruct & retail & 60.3 & 65.5 & +5.2 & [$-$0.4, +10.8] & 44.1 & 48.9 & +4.8 & [$-$2.3, +12.4] & 34.4 & 38.2 & +3.8 & [$-$4.6, +12.2] & 28.2 & 30.8 & +2.7 & [$-$6.4, +12.3] \\
7b\_r & Qwen2.5-32B-Instruct & telecom & 55.6 & 56.2 & +0.6 & [$-$4.1, +6.2] & 46.0 & 44.9 & $-$1.1 & [$-$6.1, +4.1] & 41.2 & 39.1 & $-$2.1 & [$-$8.3, +3.9] & 37.9 & 34.9 & $-$3.0 & [$-$10.3, +4.3] \\
7b\_t & Qwen2.5-32B-Instruct & airline & 26.9 & 35.0 & +8.1 & [$-$1.9, +18.8] & 12.7 & 21.4 & +8.8 & [$-$2.3, +19.3] & 6.5 & 13.6 & +7.1 & [$-$3.1, +17.2] & 3.5 & 8.6 & +5.1 & [$-$3.4, +13.4] \\
7b\_t & Qwen2.5-32B-Instruct & retail & 60.3 & 61.3 & +5.2 & [$-$2.6, +12.9] & 44.1 & 50.2 & +6.2 & [$-$3.3, +16.1] & 34.4 & 40.6 & +6.2 & [$-$4.3, +18.0] & 28.2 & 34.1 & +6.0 & [$-$5.6, +18.4] \\
7b\_t & Qwen2.5-32B-Instruct & telecom & 55.6 & 55.3 & $-$0.3 & [$-$6.9, +5.9] & 46.0 & 44.2 & $-$1.8 & [$-$9.4, +5.3] & 41.2 & 39.3 & $-$1.9 & [$-$12.0, +6.9] & 37.9 & 36.4 & $-$1.5 & [$-$13.3, +10.1] \\
14b\_a & Qwen2.5-32B-Instruct & airline & 26.9 & 31.9 & +5.0 & [$-$5.0, +15.6] & 12.7 & 19.1 & +6.4 & [$-$4.8, +18.2] & 6.5 & 12.5 & +6.0 & [$-$5.0, +16.9] & 3.5 & 8.8 & +5.3 & [$-$5.1, +17.1] \\
14b\_a & Qwen2.5-32B-Instruct & retail & 60.3 & 64.2 & +6.0 & [$-$1.3, +12.9] & 44.1 & 50.5 & +6.4 & [$-$2.6, +15.8] & 34.4 & 40.9 & +6.5 & [$-$3.5, +17.1] & 28.2 & 34.4 & +6.2 & [$-$5.2, +18.5] \\
14b\_a & Qwen2.5-32B-Instruct & telecom & 55.6 & 55.3 & $-$0.3 & [$-$5.3, +5.3] & 46.0 & 45.4 & $-$0.6 & [$-$5.3, +4.5] & 41.2 & 39.9 & $-$1.3 & [$-$5.8, +3.0] & 37.9 & 36.2 & $-$1.7 & [$-$6.9, +3.2] \\
14b\_r & Qwen2.5-32B-Instruct & airline & 26.9 & 36.9 & +10.0 & [$-$1.2, +21.9] & 12.7 & 25.0 & +12.3$\bigstar$ & [+1.2, +24.6] & 6.5 & 18.5 & +12.0$\bigstar$ & [+1.7, +23.7] & 3.5 & 13.9 & +10.4$\bigstar$ & [+0.4, +22.6] \\
14b\_r & Qwen2.5-32B-Instruct & retail & 60.3 & 71.1 & +10.8$\bigstar$ & [+3.4, +18.1] & 44.1 & 58.0 & +13.9$\bigstar$ & [+3.2, +24.5] & 34.4 & 50.5 & +16.1$\bigstar$ & [+2.8, +28.8] & 28.2 & 45.6 & +17.4$\bigstar$ & [+3.4, +32.7] \\
14b\_r & Qwen2.5-32B-Instruct & telecom & 55.6 & 53.8 & $-$1.9 & [$-$5.9, +2.2] & 46.0 & 43.2 & $-$2.8 & [$-$6.2, +0.4] & 41.2 & 38.5 & $-$2.7 & [$-$6.8, +1.2] & 37.9 & 35.4 & $-$2.5 & [$-$7.3, +2.2] \\
14b\_t & Qwen2.5-32B-Instruct & airline & 26.9 & 28.7 & +1.9 & [$-$3.1, +7.5] & 12.7 & 15.2 & +2.5 & [$-$1.2, +6.8] & 6.5 & 8.7 & +2.1 & [$-$0.5, +5.3] & 3.5 & 5.1 & +1.6 & [$-$0.1, +3.9] \\
14b\_t & Qwen2.5-32B-Instruct & retail & 60.3 & 60.0 & +1.7 & [$-$4.3, +8.2] & 44.1 & 44.5 & +0.4 & [$-$6.5, +7.4] & 34.4 & 34.4 & $-$0.0 & [$-$7.4, +7.5] & 28.2 & 28.6 & +0.4 & [$-$8.0, +9.6] \\
14b\_t & Qwen2.5-32B-Instruct & telecom & 55.6 & 58.8 & +3.1 & [$-$2.8, +8.4] & 46.0 & 48.7 & +2.7 & [$-$2.5, +8.1] & 41.2 & 44.2 & +3.0 & [$-$3.3, +10.7] & 37.9 & 41.2 & +3.4 & [$-$4.9, +11.7] \\
32b\_a & Qwen2.5-32B-Instruct & airline & 26.9 & 33.1 & +6.2 & [$-$4.4, +18.8] & 12.7 & 17.5 & +4.8 & [$-$7.0, +16.6] & 6.5 & 10.0 & +3.5 & [$-$6.1, +13.2] & 3.5 & 5.8 & +2.3 & [$-$5.1, +9.9] \\
32b\_a & Qwen2.5-32B-Instruct & retail & 60.3 & 64.2 & +3.9 & [$-$2.2, +11.2] & 44.1 & 48.0 & +3.9 & [$-$5.2, +12.8] & 34.4 & 37.5 & +3.1 & [$-$7.3, +13.2] & 28.2 & 30.4 & +2.3 & [$-$9.1, +13.6] \\
32b\_a & Qwen2.5-32B-Instruct & telecom & 55.6 & 53.4 & $-$2.2 & [$-$7.5, +3.1] & 46.0 & 43.5 & $-$2.5 & [$-$7.1, +2.2] & 41.2 & 39.7 & $-$1.5 & [$-$6.4, +3.7] & 37.9 & 37.3 & $-$0.6 & [$-$6.1, +5.9] \\
32b\_r & Qwen2.5-32B-Instruct & airline & 26.9 & 26.2 & $-$0.6 & [$-$10.0, +8.8] & 12.7 & 15.5 & +2.9 & [$-$5.5, +12.7] & 6.5 & 10.7 & +4.2 & [$-$3.6, +15.4] & 3.5 & 7.9 & +4.4 & [$-$3.0, +15.3] \\
32b\_r & Qwen2.5-32B-Instruct & retail & 60.3 & 67.2 & +6.9$\bigstar$ & [+0.9, +13.4] & 44.1 & 52.6 & +8.5$\bigstar$ & [+0.9, +16.6] & 34.4 & 42.1 & +7.7 & [$-$1.2, +16.1] & 28.2 & 34.2 & +6.1 & [$-$3.7, +15.9] \\
32b\_r & Qwen2.5-32B-Instruct & telecom & 55.6 & 53.1 & $-$2.5 & [$-$7.8, +2.5] & 46.0 & 41.9 & $-$4.1 & [$-$8.3, +0.0] & 41.2 & 36.5 & $-$4.7 & [$-$9.7, +0.2] & 37.9 & 32.9 & $-$5.0 & [$-$10.8, +0.7] \\
32b\_t & Qwen2.5-32B-Instruct & airline & 26.9 & 35.6 & +8.8 & [+0.0, +17.5] & 12.7 & 22.9 & +10.2$\bigstar$ & [+0.4, +21.6] & 6.5 & 17.1 & +10.5$\bigstar$ & [+0.2, +23.2] & 3.5 & 13.3 & +9.8 & [$-$0.1, +22.8] \\
32b\_t & Qwen2.5-32B-Instruct & retail & 60.3 & 66.4 & +6.0$\bigstar$ & [+0.4, +12.1] & 44.1 & 51.4 & +7.3 & [$-$0.6, +16.7] & 34.4 & 42.9 & +8.4 & [$-$1.6, +20.0] & 28.2 & 37.6 & +9.5 & [$-$1.7, +21.9] \\
32b\_t & Qwen2.5-32B-Instruct & telecom & 55.6 & 58.1 & +2.5 & [$-$2.2, +7.5] & 46.0 & 45.3 & $-$0.7 & [$-$6.3, +4.6] & 41.2 & 39.1 & $-$2.1 & [$-$9.6, +4.8] & 37.9 & 35.1 & $-$2.8 & [$-$11.8, +6.2] \\
\bottomrule
\end{tabular}}%
\caption{TAU2-Bench in-family playbook transfer: full Pass$^k$ ($k$=1--4) with baseline, paired $\Delta$, and CI$_{95}$ (81 routes). BL = baseline Pass$^k$; PB = playbook Pass$^k$. $\bigstar$ = paired CI$_{95}$ strictly excludes~0.}
\label{tab:app_tau2_l2_full}
\end{table*}

\begin{table*}[t]
\centering
\tiny
\setlength{\tabcolsep}{1.2pt}
\resizebox{\textwidth}{!}{%
\begin{tabular}{@{}lll r r r l r r r l r r r l r r r l@{}}
\toprule
& & & \multicolumn{4}{c}{\textbf{Pass$^1$}} & \multicolumn{4}{c}{\textbf{Pass$^2$}} & \multicolumn{4}{c}{\textbf{Pass$^3$}} & \multicolumn{4}{c}{\textbf{Pass$^4$}} \\
\cmidrule(lr){4-7} \cmidrule(lr){8-11} \cmidrule(lr){12-15} \cmidrule(lr){16-19}
\textbf{Tgt} & \textbf{Dom} & \textbf{Src} & \textbf{BL} & \textbf{PB} & \textbf{$\Delta$} & \textbf{CI$_{95}$} & \textbf{BL} & \textbf{PB} & \textbf{$\Delta$} & \textbf{CI$_{95}$} & \textbf{BL} & \textbf{PB} & \textbf{$\Delta$} & \textbf{CI$_{95}$} & \textbf{BL} & \textbf{PB} & \textbf{$\Delta$} & \textbf{CI$_{95}$} \\
\midrule
Qwen3-8B & airline & 7b\_a & 16.2 & 16.9 & +0.6 & [$-$5.0, +6.2] & 5.7 & 6.2 & +0.5 & [$-$4.1, +4.5] & 2.2 & 2.7 & +0.4 & [$-$2.4, +3.0] & 0.8 & 1.2 & +0.4 & [$-$0.9, +2.1] \\
Qwen3-8B & airline & 7b\_r & 16.2 & 11.2 & $-$5.0 & [$-$11.9, +0.6] & 5.7 & 3.4 & $-$2.3 & [$-$7.7, +2.0] & 2.2 & 1.1 & $-$1.2 & [$-$4.3, +1.5] & 0.8 & 0.4 & $-$0.4 & [$-$1.6, +0.7] \\
Qwen3-8B & airline & 7b\_t & 16.2 & 15.6 & $-$0.6 & [$-$6.9, +6.9] & 5.7 & 6.2 & +0.5 & [$-$5.4, +7.9] & 2.2 & 2.9 & +0.7 & [$-$3.2, +5.4] & 0.8 & 1.4 & +0.6 & [$-$1.4, +3.5] \\
Qwen3-8B & airline & 32b\_a & 16.2 & 19.4 & +3.1 & [$-$3.1, +10.0] & 5.7 & 8.2 & +2.5 & [$-$2.5, +8.4] & 2.2 & 3.5 & +1.2 & [$-$2.0, +5.4] & 0.8 & 1.5 & +0.7 & [$-$0.9, +3.3] \\
Qwen3-8B & airline & 32b\_r & 16.2 & 15.0 & $-$1.2 & [$-$6.2, +3.1] & 5.7 & 5.7 & +0.0 & [$-$3.2, +2.7] & 2.2 & 2.2 & $-$0.0 & [$-$2.1, +1.9] & 0.8 & 0.8 & +0.0 & [$-$1.1, +0.9] \\
Qwen3-8B & airline & 32b\_t & 16.2 & 18.8 & +2.5 & [$-$5.0, +10.6] & 5.7 & 8.9 & +3.2 & [$-$3.6, +11.2] & 2.2 & 5.4 & +3.2 & [$-$1.8, +9.9] & 0.8 & 3.6 & +2.9 & [$-$0.7, +8.3] \\
Qwen3-8B & retail & 7b\_a & 44.9 & 42.0 & $-$1.7 & [$-$8.3, +5.0] & 32.1 & 33.5 & +3.7 & [$-$5.5, +13.7] & 22.5 & 29.3 & +7.8 & [$-$3.4, +20.1] & 16.3 & 26.9 & +10.6 & [$-$0.7, +23.9] \\
Qwen3-8B & retail & 7b\_r & 44.9 & 40.9 & $-$2.8 & [$-$9.3, +3.2] & 32.1 & 29.7 & +0.6 & [$-$7.6, +9.0] & 22.5 & 24.5 & +2.9 & [$-$6.2, +12.1] & 16.3 & 20.9 & +4.6 & [$-$4.5, +14.5] \\
Qwen3-8B & retail & 7b\_t & 44.9 & 43.8 & +1.6 & [$-$5.5, +8.6] & 32.1 & 32.4 & +3.6 & [$-$5.2, +13.1] & 22.5 & 25.6 & +4.9 & [$-$6.1, +15.7] & 16.3 & 22.0 & +5.8 & [$-$6.0, +18.3] \\
Qwen3-8B & retail & 32b\_a & 44.9 & 50.4 & +2.5 & [$-$4.6, +9.2] & 32.1 & 37.8 & +5.7 & [$-$3.1, +13.4] & 22.5 & 30.9 & +8.4 & [$-$0.6, +18.7] & 16.3 & 26.8 & +10.5$\bigstar$ & [+0.9, +21.1] \\
Qwen3-8B & retail & 32b\_r & 44.9 & 39.0 & $-$3.6 & [$-$10.5, +2.8] & 32.1 & 29.1 & $-$1.0 & [$-$7.5, +5.7] & 22.5 & 23.0 & +0.5 & [$-$6.8, +8.7] & 16.3 & 17.6 & +1.3 & [$-$6.4, +10.2] \\
Qwen3-8B & retail & 32b\_t & 44.9 & 44.1 & +2.1 & [$-$2.9, +7.1] & 32.1 & 33.5 & +4.8 & [$-$1.1, +11.1] & 22.5 & 28.7 & +6.2 & [$-$0.6, +13.5] & 16.3 & 23.3 & +7.1$\bigstar$ & [+0.0, +15.6] \\
Qwen3-8B & telecom & 7b\_a & 30.0 & 35.0 & +5.0 & [$-$1.9, +11.6] & 11.2 & 15.2 & +4.0 & [$-$1.3, +9.6] & 4.8 & 8.1 & +3.3 & [$-$1.1, +8.1] & 2.1 & 5.0 & +2.9 & [$-$0.5, +7.8] \\
Qwen3-8B & telecom & 7b\_r & 30.0 & 35.3 & +5.3 & [$-$1.6, +12.2] & 11.2 & 15.5 & +4.4 & [$-$1.4, +10.1] & 4.8 & 6.9 & +2.1 & [$-$1.9, +5.8] & 2.1 & 2.9 & +0.8 & [$-$1.5, +2.9] \\
Qwen3-8B & telecom & 7b\_t & 30.0 & 35.9 & +5.9 & [$-$0.3, +12.2] & 11.2 & 15.5 & +4.4 & [$-$0.8, +9.6] & 4.8 & 7.3 & +2.5 & [$-$1.3, +6.0] & 2.1 & 3.5 & +1.4 & [$-$1.1, +3.9] \\
Qwen3-8B & telecom & 32b\_a & 30.0 & 30.9 & +0.9 & [$-$5.6, +6.9] & 11.2 & 10.7 & $-$0.4 & [$-$5.5, +4.6] & 4.8 & 3.7 & $-$1.1 & [$-$4.5, +2.2] & 2.1 & 1.1 & $-$1.0 & [$-$3.1, +0.8] \\
Qwen3-8B & telecom & 32b\_r & 30.0 & 24.7 & $-$5.3 & [$-$13.8, +2.5] & 11.2 & 9.3 & $-$1.9 & [$-$8.0, +3.5] & 4.8 & 4.0 & $-$0.8 & [$-$4.4, +2.7] & 2.1 & 1.6 & $-$0.6 & [$-$2.9, +1.4] \\
Qwen3-8B & telecom & 32b\_t & 30.0 & 20.6 & $-$9.4$\bigstar$ & [$-$14.7, $-$4.1] & 11.2 & 8.0 & $-$3.1 & [$-$7.0, +1.5] & 4.8 & 4.2 & $-$0.6 & [$-$3.8, +3.3] & 2.1 & 2.4 & +0.2 & [$-$2.1, +3.3] \\
\midrule
Qwen3-32B & airline & 7b\_a & 20.6 & 19.4 & $-$1.2 & [$-$6.9, +5.0] & 12.1 & 8.8 & $-$3.4 & [$-$8.0, +1.1] & 9.4 & 5.4 & $-$4.0$\dagger$ & [$-$8.9, +0.0] & 7.9 & 3.6 & $-$4.3$\dagger$ & [$-$10.3, +0.0] \\
Qwen3-32B & airline & 7b\_r & 20.6 & 23.8 & +3.1 & [$-$2.5, +8.8] & 12.1 & 13.9 & +1.8 & [$-$3.6, +7.7] & 9.4 & 9.7 & +0.4 & [$-$4.2, +4.9] & 7.9 & 7.5 & $-$0.4 & [$-$4.3, +2.7] \\
Qwen3-32B & airline & 7b\_t & 20.6 & 24.4 & +3.8 & [$-$2.5, +10.0] & 12.1 & 12.9 & +0.7 & [$-$6.3, +7.1] & 9.4 & 7.5 & $-$1.9 & [$-$9.4, +3.6] & 7.9 & 4.4 & $-$3.5 & [$-$11.8, +1.4] \\
Qwen3-32B & airline & 32b\_a & 20.6 & 28.1 & +7.5 & [$-$2.5, +17.5] & 12.1 & 15.5 & +3.4 & [$-$8.0, +15.0] & 9.4 & 9.8 & +0.4 & [$-$11.6, +10.8] & 7.9 & 6.1 & $-$1.8 & [$-$12.7, +8.2] \\
Qwen3-32B & airline & 32b\_r & 20.6 & 17.5 & $-$3.1 & [$-$10.0, +2.5] & 12.1 & 9.1 & $-$3.0 & [$-$11.2, +3.0] & 9.4 & 5.5 & $-$3.8 & [$-$11.8, +2.1] & 7.9 & 3.6 & $-$4.3 & [$-$12.0, +1.4] \\
Qwen3-32B & airline & 32b\_t & 20.6 & 21.2 & +0.6 & [$-$5.6, +6.2] & 12.1 & 8.9 & $-$3.2 & [$-$10.4, +2.1] & 9.4 & 4.1 & $-$5.3 & [$-$13.7, +0.7] & 7.9 & 1.9 & $-$6.1 & [$-$16.1, +0.1] \\
Qwen3-32B & retail & 7b\_a & 72.4 & 60.8 & $-$9.5$\bigstar$ & [$-$14.7, $-$4.3] & 60.2 & 51.6 & $-$8.6$\bigstar$ & [$-$15.0, $-$1.6] & 52.8 & 45.3 & $-$7.5 & [$-$15.6, +1.5] & 47.3 & 40.5 & $-$6.8 & [$-$16.3, +2.8] \\
Qwen3-32B & retail & 7b\_r & 72.4 & 59.7 & $-$8.6$\bigstar$ & [$-$15.5, $-$2.6] & 60.2 & 51.4 & $-$8.9$\bigstar$ & [$-$17.6, $-$0.7] & 52.8 & 43.0 & $-$9.7 & [$-$20.1, +0.6] & 47.3 & 36.3 & $-$11.0 & [$-$23.4, +0.8] \\
Qwen3-32B & retail & 7b\_t & 72.4 & 65.4 & $-$4.7 & [$-$11.2, +1.7] & 60.2 & 56.9 & $-$3.3 & [$-$11.5, +4.9] & 52.8 & 49.8 & $-$3.0 & [$-$13.1, +7.9] & 47.3 & 44.6 & $-$2.7 & [$-$15.0, +9.6] \\
Qwen3-32B & retail & 32b\_a & 72.4 & 64.6 & $-$5.6$\bigstar$ & [$-$11.2, $-$0.4] & 60.2 & 55.9 & $-$4.3 & [$-$11.9, +3.3] & 52.8 & 49.9 & $-$2.8 & [$-$12.3, +6.2] & 47.3 & 45.6 & $-$1.7 & [$-$12.5, +8.9] \\
Qwen3-32B & retail & 32b\_r & 72.4 & 64.6 & $-$5.6 & [$-$12.5, +0.9] & 60.2 & 53.8 & $-$6.4 & [$-$15.8, +3.1] & 52.8 & 46.1 & $-$6.7 & [$-$19.1, +4.0] & 47.3 & 40.4 & $-$6.8 & [$-$20.4, +5.4] \\
Qwen3-32B & retail & 32b\_t & 72.4 & 62.5 & $-$7.8$\bigstar$ & [$-$13.8, $-$2.2] & 60.2 & 53.1 & $-$7.1$\dagger$ & [$-$14.7, +0.0] & 52.8 & 45.6 & $-$7.1 & [$-$16.5, +2.0] & 47.3 & 40.3 & $-$7.0 & [$-$18.2, +3.6] \\
Qwen3-32B & telecom & 7b\_a & 40.9 & 47.8 & +6.9$\bigstar$ & [+0.9, +12.8] & 28.6 & 32.8 & +4.2 & [$-$2.9, +11.2] & 21.5 & 25.2 & +3.7 & [$-$3.6, +11.7] & 16.5 & 20.5 & +4.0 & [$-$4.4, +13.0] \\
Qwen3-32B & telecom & 7b\_r & 40.9 & 47.8 & +6.9$\bigstar$ & [+0.9, +12.8] & 28.6 & 34.3 & +5.7 & [$-$2.4, +13.8] & 21.5 & 27.5 & +6.1 & [$-$3.6, +16.2] & 16.5 & 23.1 & +6.6 & [$-$3.8, +17.3] \\
Qwen3-32B & telecom & 7b\_t & 40.9 & 48.8 & +7.8$\bigstar$ & [+1.2, +14.4] & 28.6 & 33.4 & +4.8 & [$-$2.5, +12.0] & 21.5 & 25.0 & +3.5 & [$-$4.9, +11.6] & 16.5 & 19.5 & +3.0 & [$-$7.0, +12.0] \\
Qwen3-32B & telecom & 32b\_a & 40.9 & 45.3 & +4.4 & [$-$0.3, +9.1] & 28.6 & 31.2 & +2.7 & [$-$2.9, +8.0] & 21.5 & 23.3 & +1.8 & [$-$4.5, +8.4] & 16.5 & 17.6 & +1.1 & [$-$6.2, +8.8] \\
Qwen3-32B & telecom & 32b\_r & 40.9 & 45.3 & +4.4$\dagger$ & [+0.0, +9.1] & 28.6 & 29.8 & +1.2 & [$-$3.4, +6.5] & 21.5 & 22.2 & +0.7 & [$-$4.7, +6.4] & 16.5 & 17.1 & +0.6 & [$-$5.1, +6.5] \\
Qwen3-32B & telecom & 32b\_t & 40.9 & 40.6 & $-$0.3 & [$-$6.2, +5.6] & 28.6 & 24.3 & $-$4.3 & [$-$12.0, +2.6] & 21.5 & 15.8 & $-$5.6 & [$-$13.6, +1.8] & 16.5 & 10.9 & $-$5.7 & [$-$14.8, +2.6] \\
\midrule
Llama3.1-70B-Inst & airline & 7b\_a & 26.2 & 26.2 & +0.0 & [$-$3.8, +3.8] & 22.5 & 20.9 & $-$1.6 & [$-$7.9, +3.9] & 20.3 & 17.6 & $-$2.7 & [$-$10.9, +3.0] & 18.6 & 15.0 & $-$3.6 & [$-$13.2, +2.0] \\
Llama3.1-70B-Inst & airline & 7b\_r & 26.2 & 26.9 & +0.6 & [$-$3.1, +5.6] & 22.5 & 22.7 & +0.2 & [$-$6.8, +7.1] & 20.3 & 20.3 & +0.0 & [$-$9.6, +9.6] & 18.6 & 18.6 & +0.0 & [$-$11.8, +11.8] \\
Llama3.1-70B-Inst & airline & 7b\_t & 26.2 & 28.1 & +1.9 & [$-$1.2, +5.0] & 22.5 & 23.0 & +0.5 & [$-$3.6, +5.0] & 20.3 & 19.8 & $-$0.4 & [$-$5.6, +4.3] & 18.6 & 17.1 & $-$1.5 & [$-$7.5, +3.0] \\
Llama3.1-70B-Inst & airline & 32b\_a & 26.2 & 30.0 & +3.8$\dagger$ & [+0.0, +9.4] & 22.5 & 27.7 & +5.2$\dagger$ & [+0.0, +14.3] & 20.3 & 26.8 & +6.5$\dagger$ & [+0.0, +17.7] & 18.6 & 26.1 & +7.4$\dagger$ & [+0.0, +19.7] \\
Llama3.1-70B-Inst & airline & 32b\_r & 26.2 & 26.2 & +0.0 & [$-$4.4, +4.4] & 22.5 & 21.2 & $-$1.2 & [$-$6.6, +3.6] & 20.3 & 18.0 & $-$2.2 & [$-$9.2, +3.8] & 18.6 & 15.7 & $-$2.9 & [$-$10.1, +3.1] \\
Llama3.1-70B-Inst & airline & 32b\_t & 26.2 & 28.1 & +1.9 & [$-$1.2, +6.2] & 22.5 & 25.5 & +3.0 & [$-$1.8, +9.5] & 20.3 & 24.0 & +3.8 & [$-$1.8, +11.2] & 18.6 & 22.9 & +4.2 & [$-$1.4, +12.2] \\
Llama3.1-70B-Inst & retail & 7b\_a & 8.2 & 7.8 & +0.4$\dagger$ & [+0.0, +1.3] & 7.0 & 7.0 & +0.2$\dagger$ & [+0.0, +0.7] & 6.9 & 7.0 & +0.1$\dagger$ & [+0.0, +0.2] & 6.9 & 6.9 & +0.0$\dagger$ & [+0.0, +0.0] \\
Llama3.1-70B-Inst & retail & 7b\_r & 8.2 & 8.5 & +0.9$\dagger$ & [+0.0, +2.2] & 7.0 & 7.4 & +0.4$\dagger$ & [+0.0, +1.0] & 6.9 & 7.0 & +0.1$\dagger$ & [+0.0, +0.2] & 6.9 & 6.9 & +0.0$\dagger$ & [+0.0, +0.0] \\
Llama3.1-70B-Inst & retail & 7b\_t & 8.2 & 8.6 & +0.4 & [$-$1.3, +2.6] & 7.0 & 6.8 & $-$0.2 & [$-$2.6, +1.8] & 6.9 & 5.8 & $-$1.0 & [$-$3.9, +0.7] & 6.9 & 5.2 & $-$1.7 & [$-$5.2, +0.1] \\
Llama3.1-70B-Inst & retail & 32b\_a & 8.2 & 9.1 & +0.9 & [$-$1.3, +3.9] & 7.0 & 8.1 & +1.1$\dagger$ & [+0.0, +3.3] & 6.9 & 7.5 & +0.6$\dagger$ & [+0.0, +1.8] & 6.9 & 7.1 & +0.2$\dagger$ & [+0.0, +0.7] \\
Llama3.1-70B-Inst & retail & 32b\_r & 8.2 & 9.3 & +1.7 & [$-$1.3, +6.5] & 7.0 & 9.2 & +2.5$\dagger$ & [+0.0, +7.4] & 6.9 & 9.1 & +2.2$\dagger$ & [+0.0, +6.5] & 6.9 & 8.6 & +1.7$\dagger$ & [+0.0, +5.2] \\
Llama3.1-70B-Inst & retail & 32b\_t & 8.2 & 8.8 & +0.9$\dagger$ & [+0.0, +2.6] & 7.0 & 7.4 & +0.6$\dagger$ & [+0.0, +1.8] & 6.9 & 7.1 & +0.2$\dagger$ & [+0.0, +0.7] & 6.9 & 6.9 & +0.0$\dagger$ & [+0.0, +0.1] \\
Llama3.1-70B-Inst & telecom & 7b\_a & 23.1 & 26.9 & +3.8 & [$-$0.9, +8.1] & 11.6 & 14.3 & +2.7 & [$-$1.7, +7.3] & 5.8 & 8.8 & +3.0 & [$-$0.8, +7.3] & 2.8 & 5.6 & +2.7 & [$-$0.2, +6.2] \\
Llama3.1-70B-Inst & telecom & 7b\_r & 23.1 & 24.1 & +0.9 & [$-$3.8, +5.3] & 11.6 & 11.9 & +0.3 & [$-$3.3, +3.6] & 5.8 & 6.2 & +0.3 & [$-$2.5, +2.9] & 2.8 & 3.0 & +0.2 & [$-$1.7, +1.9] \\
Llama3.1-70B-Inst & telecom & 7b\_t & 23.1 & 24.4 & +1.2 & [$-$5.0, +6.6] & 11.6 & 11.5 & $-$0.1 & [$-$5.0, +4.6] & 5.8 & 6.0 & +0.1 & [$-$3.8, +4.5] & 2.8 & 3.2 & +0.4 & [$-$2.5, +3.6] \\
Llama3.1-70B-Inst & telecom & 32b\_a & 23.1 & 25.0 & +1.9 & [$-$2.8, +6.6] & 11.6 & 13.8 & +2.2 & [$-$2.7, +7.5] & 5.8 & 8.7 & +2.9 & [$-$1.5, +7.6] & 2.8 & 5.7 & +2.9 & [$-$0.9, +6.9] \\
Llama3.1-70B-Inst & telecom & 32b\_r & 23.1 & 28.4 & +5.3$\bigstar$ & [+1.9, +8.8] & 11.6 & 16.0 & +4.4$\bigstar$ & [+0.8, +8.4] & 5.8 & 9.9 & +4.1$\bigstar$ & [+0.6, +8.4] & 2.8 & 6.1 & +3.3$\bigstar$ & [+0.4, +7.1] \\
Llama3.1-70B-Inst & telecom & 32b\_t & 23.1 & 23.8 & +0.6 & [$-$4.4, +5.6] & 11.6 & 11.1 & $-$0.5 & [$-$5.0, +4.3] & 5.8 & 6.1 & +0.2 & [$-$4.0, +4.7] & 2.8 & 3.6 & +0.8 & [$-$2.5, +4.9] \\
\bottomrule
\end{tabular}}%
\caption{TAU2-Bench cross-family playbook transfer: full Pass$^k$ ($k$=1--4) with baseline, paired $\Delta$, and CI$_{95}$ (54 routes). $\bigstar$ = CI$_{95}$ strictly excludes 0; $\dagger$ = boundary.}
\label{tab:app_tau2_cross_l2_full}
\end{table*}

\subsection{Paired-CI Statistical Analysis}
\label{sec:app_tau2_heatmaps}

We run eight submitted trials per task (2$\times$ the official default) and compute paired bootstrap CIs on the task intersection. For the presubmission headline metric Pass$^1$, two-sided centered paired-task bootstrap $p$-values use 500,000 resamples. We apply Holm correction jointly over all 135 route hypotheses as the primary family-wise analysis, and report 81-route in-family and 54-route cross-family Holm corrections plus global BH as sensitivity analyses. Pass$^{2\text{--}4}$, costs, and routes not retained by correction are descriptive. Full per-configuration point estimates and intervals remain useful for direction, magnitude, and uncertainty, but are not confirmatory single-cell labels.

\begin{figure*}[t]
\centering
\includegraphics[width=\textwidth]{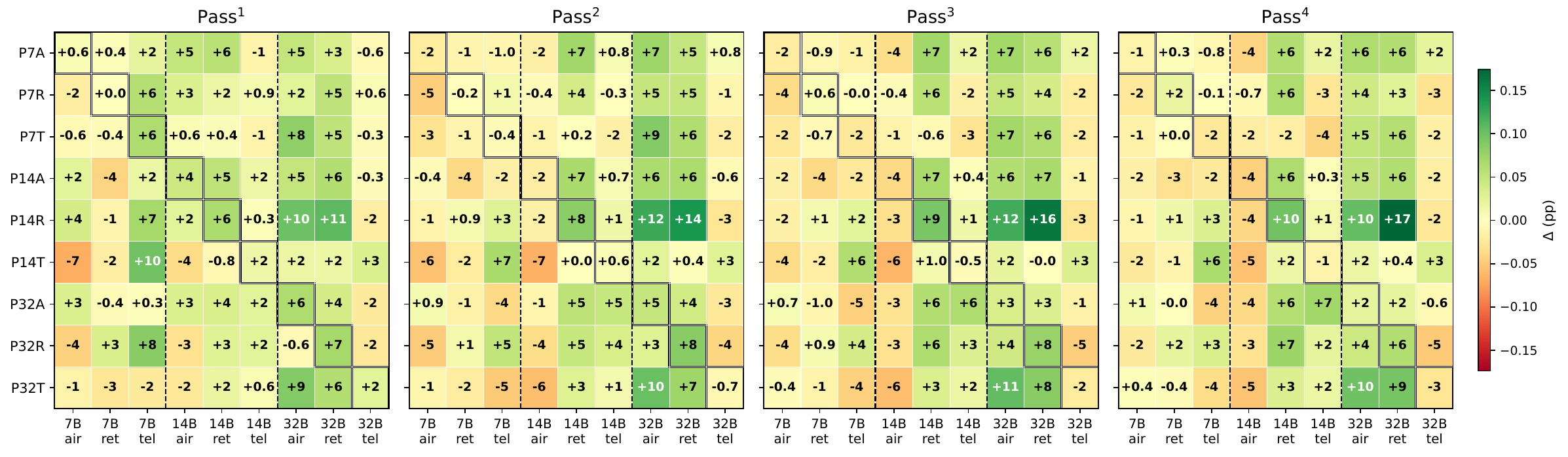}
\caption{In-family (Qwen2.5) $\Delta$\,Pass$^k$ heatmaps ($k=1,\ldots,4$, 81 configurations). Dashed lines separate target-size groups; black outlines mark same-size-same-domain diagonal cells. Source playbooks use the P-prefix scheme (e.g.\ P14R = Qwen2.5-14B-Instruct retail playbook).}
\label{fig:tau2_infamily_heatmap}
\end{figure*}

\begin{figure*}[t]
\centering
\includegraphics[width=\textwidth]{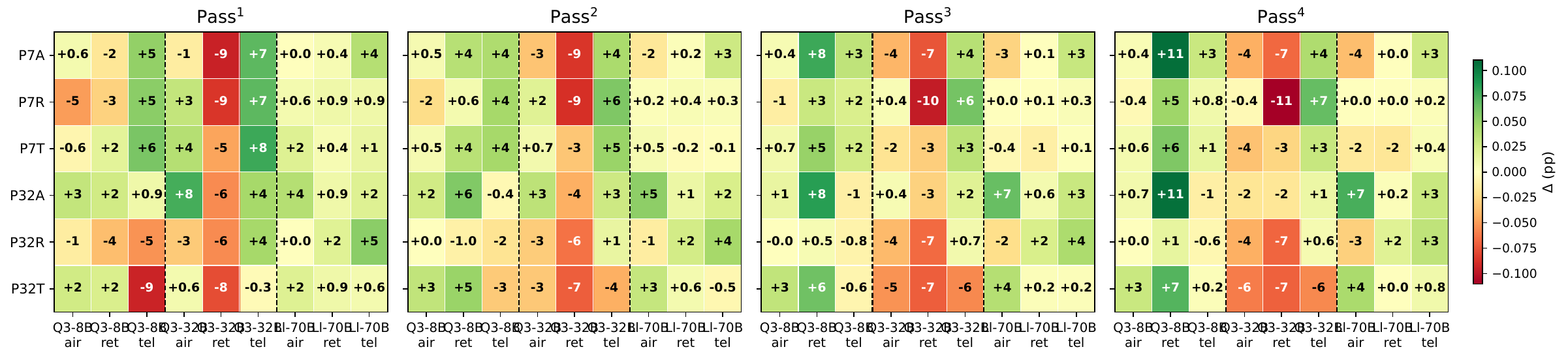}
\caption{Cross-family $\Delta$\,Pass$^k$ heatmaps ($k=1,\ldots,4$, 54 configurations). Targets: Qwen3-8B, Qwen3-32B, Llama3.1-70B-Inst. Dashed lines separate target groups.}
\label{fig:tau2_crossfam_heatmap}
\end{figure*}

\begin{figure*}[t]
\centering
\includegraphics[width=0.47\textwidth]{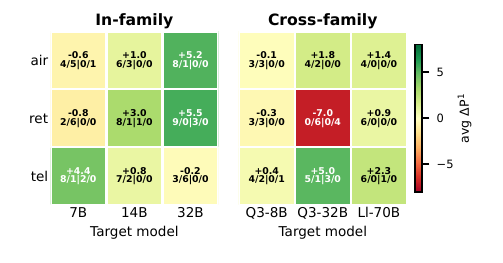}%
\hfill
\includegraphics[width=0.47\textwidth]{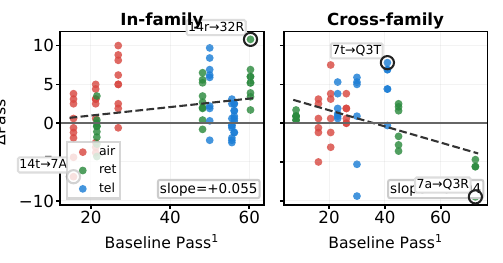}
\caption{Descriptive TAU2-Bench summary views. Left: target-condition averages and sign counts. Right: baseline Pass$^1$ versus $\Delta$Pass$^1$.}
\label{fig:tau2_summary_combined}
\end{figure*}

\subsection{Matched-vs.-Mismatched Domain Control}
\label{sec:app_tau2_domain_control}

\begin{figure}[t]
\centering
\includegraphics[width=\columnwidth]{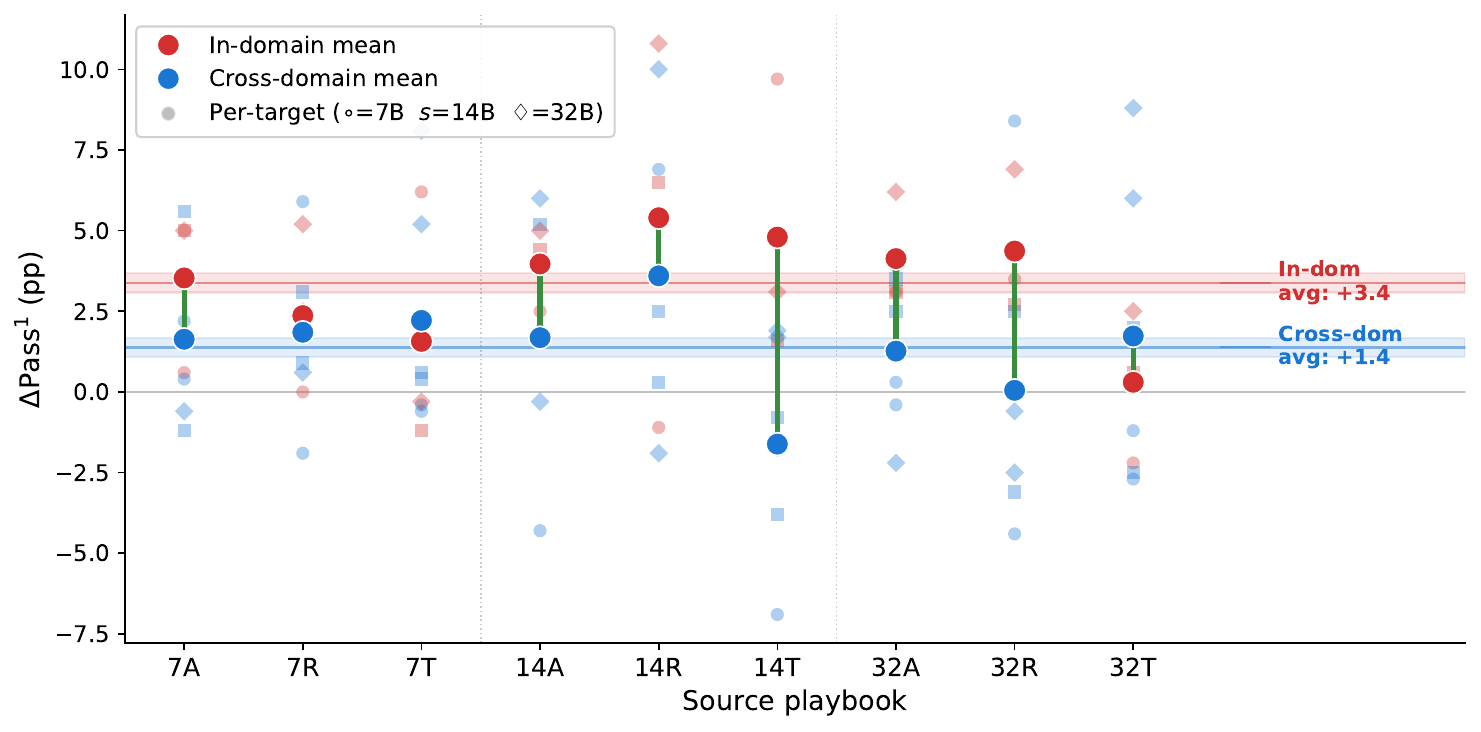}
\caption{Matched-vs.-mismatched domain control for the 81 in-family TAU2-Bench configurations. Each source playbook is applied to its matched domain and two mismatched domains; in 7 of 9 sources, the matched-domain mean is higher.}
\label{fig:app_tau2_domain_control}
\end{figure}

The in-family grid also provides a zero-extra-run control against a pure prompt-injection explanation. For each source artifact, we compare the same injected playbook file on its matched service domain and on the two mismatched domains, so prompt length and format are exactly controlled within-source. Figure~\ref{fig:app_tau2_domain_control} shows the per-source matched-vs.-mismatched comparison, and Table~\ref{tab:app_tau2_domain_control_stats} gives the corresponding aggregated test.

For each source artifact $s$, let $M_s$ denote the mean $\Delta$Pass$^1$ over the three matched-domain target conditions and $X_s$ the mean over the six mismatched-domain target conditions; we then define the within-source contrast $\Delta_s = M_s - X_s$. The reported aggregate effect is the source-paired mean $\bar{\Delta} = \mathrm{mean}_s \Delta_s$. CI$_{95}$ is a source-level bootstrap percentile interval over the nine source artifacts. Our primary $p$-value comes from an exact blocked permutation test that, within each source, reassigns the matched-domain label over the three candidate domains ($3^9 = 19683$ assignments), which matches the design and avoids a normality assumption. Paired $t$-test and Wilcoxon signed-rank results are included only as robustness checks.

\begin{table}[t]
\centering
\small
\begin{tabular}{@{}lr@{}}
\toprule
\textbf{Statistic} & \textbf{Value} \\
\midrule
Matched-domain mean $\Delta$Pass$^1$ & +3.34 pp \\
Mismatched-domain mean $\Delta$Pass$^1$ & +1.62 pp \\
Source-paired mean difference $\bar{\Delta}$ & +1.72 pp \\
Bootstrap CI$_{95}$ for $\bar{\Delta}$ & [$+0.25$, $+2.97$] pp \\
Exact blocked-permutation $p$ & 0.035 \\
Paired effect size $d_z$ & 0.77 \\
\midrule
Paired $t$-test $p$ (one-sided) & 0.025 \\
Wilcoxon signed-rank $p$ (one-sided) & 0.037 \\
\bottomrule
\end{tabular}
\caption{Matched-vs.-mismatched domain control aggregated at the source-playbook level. The primary inferential quantities are the source-paired mean difference, bootstrap CI$_{95}$, exact blocked-permutation $p$-value, and paired effect size; the bottom two lines are robustness checks.}
\label{tab:app_tau2_domain_control_stats}
\end{table}

\subsection{Target-Derived versus Transferred Playbooks}
\label{sec:app_tau2_redistill}

We re-slice an existing pre-submission L1 source-model analysis for a Qwen2.5-14B target. Within each domain, the no-playbook baseline, target-derived 14B artifact, and transferred 7B- and 32B-source artifacts share the test protocol and common valid task set. All artifacts are constructed from same-domain training trajectories. The analysis is exploratory: it was not part of the 135-route L2 multiplicity family and no new multiplicity correction is attached to these direct comparisons.

\begin{table}[t]
\centering
\scriptsize
\setlength{\tabcolsep}{3.2pt}
\begin{tabular}{@{}lrr@{}}
\toprule
Domain & Target $-$ 7B source & Target $-$ 32B source \\
\midrule
Airline & $-4.38$ [$-16.88$,$+8.13$] & $+10.00$ [$+2.50$,$+18.75$] \\
Retail & $+6.67$ [$+0.42$,$+12.92$] & $-0.42$ [$-6.67$,$+6.25$] \\
Telecom & $-0.94$ [$-5.31$,$+3.44$] & $-3.75$ [$-9.69$,$+1.88$] \\
\bottomrule
\end{tabular}
\caption{Direct paired Pass$^1$ differences (pp, CI$_{95}$); positive values favor target-side redistillation. These secondary estimands are reported without confirmatory labels.}
\label{tab:app_tau2_redistill_ci}
\end{table}

Mean tool calls remain similar within each domain. For baseline, target-derived, 7B-source, and 32B-source arms, respectively, the means are 9.35/8.88/9.14/8.41 in airline, 8.44/8.42/8.47/8.54 in retail, and 22.86/24.67/23.57/24.09 in telecom. The full three-domain pattern shows that target-side redistillation is not uniformly superior, while frozen transfer is not uniformly preferable either.

\subsection{Illustrative TAU2-Bench Cases}

These cases are illustrative rather than statistical. We select one clearly positive and one clearly negative configuration to make the compatibility story concrete at the task level.

\begin{table*}[t]
\centering
\small
\setlength{\tabcolsep}{4pt}
\begin{tabular}{@{}llllp{0.28\textwidth}p{0.22\textwidth}@{}}
\toprule
\textbf{Configuration} & \textbf{Task} & \textbf{Base} & \textbf{PB} & \textbf{Dominant failure pattern} & \textbf{Mechanism} \\
\midrule
P14R $\to$ Qwen2.5-14B-Instruct/ret & 17 & 56\% & 88\% & modify called before prerequisite read & prerequisite-read correction \\
P14T $\to$ Qwen2.5-7B-Instruct/air  & 2  & 62\% & 12\% & complaint template injects \texttt{send\_certificate} & cross-domain action-template conflict \\
\bottomrule
\end{tabular}
\caption{Illustrative TAU2-Bench cases. Percentages are task-level Pass$^1$ estimates over the task's 8 trials. These examples are chosen to expose the mechanisms behind a positive and a negative configuration, not as additional statistical evidence.}
\label{tab:app_tau2_cases}
\end{table*}

\paragraph{Positive configuration: P14R $\to$ Qwen2.5-14B-Instruct/retail, Task 17.} The user requests an address update for an existing order. In failing baseline trials, the agent often calls \texttt{modify\_pending\_order\_address} directly with placeholder fields such as the current street or city, without first retrieving the order state. The playbook corrects this by inserting the prerequisite \texttt{get\_order\_details} call, after which the agent can preserve unchanged fields and modify only the requested address slot. This is a clean same-domain gain: the imported procedure matches both the target tool semantics and the target domain's latent failure pattern.

\paragraph{Negative configuration: P14T $\to$ Qwen2.5-7B-Instruct/airline, Task 2.} The user begins with a booking request and then pivots to a complaint about a delayed flight. In successful baseline trials, the agent typically remains read-only, using actions such as \texttt{get\_user}, \texttt{get\_reservation}, and \texttt{get\_flight\_status} without taking an unnecessary compensatory action. Under the telecom-derived playbook, the agent frequently injects a \texttt{send\_certificate} step that is appropriate in telecom complaint handling but harmful in this airline task; once that action is taken, the interaction usually ends with an incorrect database state and reward 0. This case shows that negative transfer can arise from an imported action template, not just from missing knowledge.

\subsection{Same-Length Generic-Playbook Control}
\label{sec:app_tau2_placebo}

To probe whether TAU2-Bench transfer effects can be explained by prompt length or checklist-like structure alone, we evaluate a same-length generic placebo on one strong positive route and two negative routes, including a repeated cross-family failure condition. Each placebo is a route-specific, structure-matched rewrite of the corresponding source playbook into generic workflow advice, generated with Claude Opus 4.6 via Claude Code. Concretely, we preserve the section count, approximate length, and checklist-style formatting of the source artifact, but replace domain-specific procedures, tool names, prerequisite logic, and action-ordering rules with non-domain operational reminders. We keep this control small and route-specific: it is meant to weaken a pure format-driven alternative explanation, not to serve as a benchmark-wide causal intervention.

The main-text results in Table~\ref{tab:tau2_placebo_main} summarize the three routes. On P14R$\to$Qwen2.5-14B-Instruct/retail, the transferred playbook raises Pass$^1$ from 55.2\% to 62.6\%, while the placebo reaches 57.1\%. On P14T$\to$Qwen2.5-7B-Instruct/airline, the transferred playbook lowers Pass$^1$ from 15.6\% to 8.7\%, while the placebo stays near baseline at 14.4\%. On P7A$\to$Qwen3-32B/retail, the transferred playbook lowers Pass$^1$ from 72.4\% to 60.8\%, while the placebo reaches 70.7\%. This selected control weakens a pure prompt-length or checklist-structure explanation, but it is descriptive and does not establish format invariance.

\section{XBench-DeepSearch Audit}
\label{sec:app_xbench}

This section keeps the XBench evidence explicitly exploratory: we summarize the four deployment conditions, target-derived and cost-screening comparisons, the aggregate runtime-shift behavioral audit, and illustrative failure cases.

\subsection{Descriptive XBench Playbook Portability Summary}

Because XBench is a small exploratory split, we summarize the four-condition transferred-playbook comparison descriptively rather than as a benchmark-wide inferential table. The playbook itself is distilled from Tongyi-DeepResearch-30B-A3B 32K development-split trajectories, then held fixed during transfer. The pattern mirrors the main-text reading: under same-context transfer, Pass@1 improves and cost stays bounded; under cross-context transfer, Pass@3 degrades and cost becomes less portable.

\begin{table}[t]
\centering
\small
\setlength{\tabcolsep}{4pt}
\resizebox{\columnwidth}{!}{%
\begin{tabular}{@{}lrr@{}}
\toprule
\textbf{Condition} & \textbf{$\Delta$Pass@1} & \textbf{$\Delta$Pass@3} \\
\midrule
Tongyi-DeepResearch-30B-A3B 32K  & +4  & $-$2  \\
Qwen3-32B 32K  & +8  & \phantom{$-$}0  \\
Tongyi-DeepResearch-30B-A3B 128K & \phantom{+}0  & $-$16 \\
Qwen3-32B 128K & +6  & $-$6  \\
\bottomrule
\end{tabular}
}
\caption{Exploratory portability profile of the transferred XBench playbook on the XBench-DeepSearch held-out evaluation slice (\%). Same-context transfer improves Pass@1 more reliably than cross-context transfer preserves Pass@3.}
\label{tab:app_xbench_pbv1_perf}
\end{table}

\subsection{Cost-Side Descriptive Summary}

\begin{table}[t]
\centering
\small
\setlength{\tabcolsep}{3.5pt}
\resizebox{\columnwidth}{!}{%
\begin{tabular}{@{}lrrll@{}}
\toprule
\textbf{Comp.} & \textbf{$\Delta$Search} & \textbf{$\Delta$Tool} & \textbf{$\Delta$Overlong} & \textbf{Read.} \\
\midrule
Tongyi-DeepResearch-30B-A3B 32K  & $-$3\%  & $-$6\%   & 3$\to$1\%   & cost-neutral \\
Qwen3-32B 32K  & +15\%   & +12\%    & 3$\to$1\%   & mild+ \\
Tongyi-DeepResearch-30B-A3B 128K & +17\%   & +9\%     & 24$\to$29\% & worse \\
Qwen3-32B 128K & +167\%  & +136\%   & 8$\to$25\%  & severe \\
\bottomrule
\end{tabular}}
\caption{Descriptive XBench cost shifts under the transferred playbook using the three main behavioral indicators: Avg Search Calls, Avg Tool Calls, and Overlong-Trajectory Rate ($>$30 Turns).}
\label{tab:app_xbench_cost}
\end{table}

\subsection{Target-Derived versus Transferred Artifact}
\label{sec:app_xbench_redistill}

On the Qwen3-32B 128K held-out split, we compare the no-playbook baseline, transferred Tongyi-derived artifact, and an artifact constructed only from Qwen3 development-split failures and frozen before held-out evaluation. All arms contain 50 questions and three trials per question.

\begin{table}[t]
\centering
\small
\setlength{\tabcolsep}{3.5pt}
\begin{tabular}{@{}lrrr@{}}
\toprule
Arm & Pass@1 & True Pass@3 & Avg. total tools \\
\midrule
Baseline & 45.3 & 66 & 10.41 \\
Transferred PBV1 & 46.7 & 70 & 24.67 \\
Target-derived & 46.0 & 66 & 36.15 \\
\bottomrule
\end{tabular}
\caption{Exploratory Qwen3-32B 128K comparison (accuracy in \%). Paired accuracy intervals versus baseline cross zero. About 24\% of target-derived trajectories reach a token or call budget.}
\label{tab:app_xbench_redistill}
\end{table}

The target-derived artifact increases average total tool calls by 247.3\% over baseline without improving True Pass@3. This does not show that transfer is superior; it shows that target-side redistillation does not automatically produce a safer or more effective deployment artifact.

\subsection{Exploratory and Prospective Cost Screening}
\label{sec:app_xbench_precheck}

The original four-condition development analysis combines Pass@1 direction and absolute tool-call inflation into retrospective labels. Development labels agree with the eventual qualitative readout, but two 128K probes post-date evaluation, the labels have no independent ground truth, and development Pass@1 does not reliably track True Pass@3. We therefore treat this table as hypothesis generation rather than validation of a routing rule.

\begin{table}[t]
\centering
\small
\setlength{\tabcolsep}{3.2pt}
\resizebox{\columnwidth}{!}{%
\begin{tabular}{@{}lrrll@{}}
\toprule
\textbf{Condition} & \textbf{Dev $\Delta$Pass@1} & \textbf{Dev $\Delta$TC} & \textbf{Label} & \textbf{Held-out readout} \\
\midrule
Tongyi-DeepResearch-30B-A3B 32K  & +4.0   & $-$4.0  & Deploy  & Pass@1$\uparrow$, cost bounded \\
Qwen3-32B 32K  & +10.0  & +3.4    & Deploy  & Pass@1$\uparrow$, mild cost tradeoff \\
Tongyi-DeepResearch-30B-A3B 128K & $-$14.9 & $-$5.0  & Reject  & P@3$\downarrow$; not portable \\
Qwen3-32B 128K & +6.0   & +13.5   & Guarded & Pass@1$\uparrow$, but cost/Pass@3 worsen \\
\bottomrule
\end{tabular}}
\caption{Retrospective XBench development-slice labels. Their agreement with the qualitative readout is not prospective validation or independent ground truth.}
\label{tab:app_xbench_precheck}
\end{table}

During review, we additionally freeze a $+5$ mean-total-tool-call alert threshold before evaluating a new Gemini-2.5-Pro PBV1 held-out pairing. Smoke and held-out deltas are $+0.26$ and $+4.86$ calls, so both receive \emph{no alert}. The held-out result is nevertheless only 0.14 calls below threshold and is a 347.1\% relative increase. This one threshold-specific match neither calibrates the threshold nor validates cost safety, accuracy prediction, termination safety, or a general compatibility selector.

\subsection{Behavioral Audit of Re-querying and Delayed Stopping}
\label{sec:app_xbench_mech}

For Qwen3-32B, the most informative XBench failure is not a simple accuracy drop but a change in search dynamics after a 32K-calibrated playbook is moved to 128K execution. Under 32K, the transferred artifact slightly improves Pass@1 while keeping search depth controlled. Under 128K, it preserves aggressive search and verification but produces persistent re-querying and delayed submission, leading to trajectory inflation without a corresponding accuracy gain.

The absolute held-out means behind the main-text summary are as follows. For Qwen3-32B 128K, Avg Search Calls rise from 5.9 to 15.7, Avg Tool Calls from 10.4 to 24.7, and Overlong-Trajectory Rate ($>$30 turns) from 8\% to 25\%. These are the absolute values corresponding to the relative shifts in Table~\ref{tab:app_xbench_cost}.

\begin{table}[t]
\centering
\small
\setlength{\tabcolsep}{3.0pt}
\begin{tabular}{@{}lrrr@{}}
\toprule
128K trajectory property & Base & PBV1 & $\Delta$ (CI$_{95}$) \\
\midrule
Any identical query & 17.3 & 39.3 & $+22.0$ [$+12$,$+32$] \\
$\geq$3 identical searches & 8.0 & 24.0 & $+16.0$ [$+8$,$+24.7$] \\
Submit within 10 tools & 78.0 & 30.0 & $-48.0$ [$-58$,$-38$] \\
\bottomrule
\end{tabular}
\caption{Qwen3-32B 128K behavioral audit over 150 paired trajectories (\%). Intervals use question-cluster bootstrap resampling. Corresponding 32K changes are smaller and cross zero.}
\label{tab:app_xbench_behavior}
\end{table}

\begin{table}[t]
\centering
\small
\setlength{\tabcolsep}{4pt}
\resizebox{\columnwidth}{!}{%
\begin{tabular}{@{}lcccc@{}}
\toprule
\textbf{Threshold} & \textbf{32K BL} & \textbf{32K + PB} & \textbf{128K BL} & \textbf{128K + PB} \\
\midrule
$>$30 turns      & 3\%  & 1\%  & 12\% & 26\% \\
$>$20 tool calls & 6\%  & 11\% & 10\% & 42\% \\
$>$40 tool calls & 1\%  & 0\%  & 7\%  & 19\% \\
\bottomrule
\end{tabular}
}
\caption{Qwen3-32B search inflation under runtime shift. The transferred XBench playbook remains well-behaved under 32K, but under 128K it produces a much heavier tail in Overlong-Trajectory Rate and repeated tool use.}
\label{tab:app_xbench_mech_dist}
\end{table}

Table~\ref{tab:app_xbench_mech_dist} shows the shift directly. Under 32K, the transferred artifact does not create a heavy tail of long trajectories. Under 128K, the same artifact sharply increases overlong runs. Together with Table~\ref{tab:app_xbench_behavior}, this supports persistent re-querying and delayed stopping as a proximal behavioral pathway. We do not measure stopping-probability calibration, attention or position effects, or other internal context-length-OOD variables, and make no internal causal claim.

\begin{table}[t]
\centering
\small
\setlength{\tabcolsep}{4pt}
\resizebox{\columnwidth}{!}{%
\begin{tabular}{@{}lp{0.16\columnwidth}p{0.16\columnwidth}p{0.36\columnwidth}@{}}
\toprule
\textbf{Condition} & \textbf{Turns} & \textbf{Tool calls} & \textbf{Behavior} \\
\midrule
32K baseline & 26 & 25 & repeated search, then submits a wrong answer \\
32K + playbook & 4 & 3 & finds the key entities quickly and submits immediately \\
128K baseline & 22 & 7 & a few searches, then gives a bounded best-effort answer \\
128K + playbook & 100 & 100 & keeps issuing near-duplicate search variants without stopping \\
\bottomrule
\end{tabular}
}
\caption{Representative Qwen3-32B runtime-shift case: Olympic sisters age-gap question. The 128K playbook condition fails by persistent re-querying rather than by immediate factual confusion.}
\label{tab:app_xbench_mech_case}
\end{table}

The Olympic sisters age-gap question in Table~\ref{tab:app_xbench_mech_case} illustrates the aggregate behavioral pathway. On 32K, the artifact combines targeted search with early submission. On 128K, it re-queries surface variants of the same subproblem until the trajectory budget is exhausted. The case is illustrative and does not identify an internal causal mechanism.

\subsection{Illustrative XBench Case}

The following case is illustrative rather than statistical. We include it because it makes the runtime-shift failure mode directly observable.

\begin{figure}[t]
\centering
\includegraphics[width=\columnwidth]{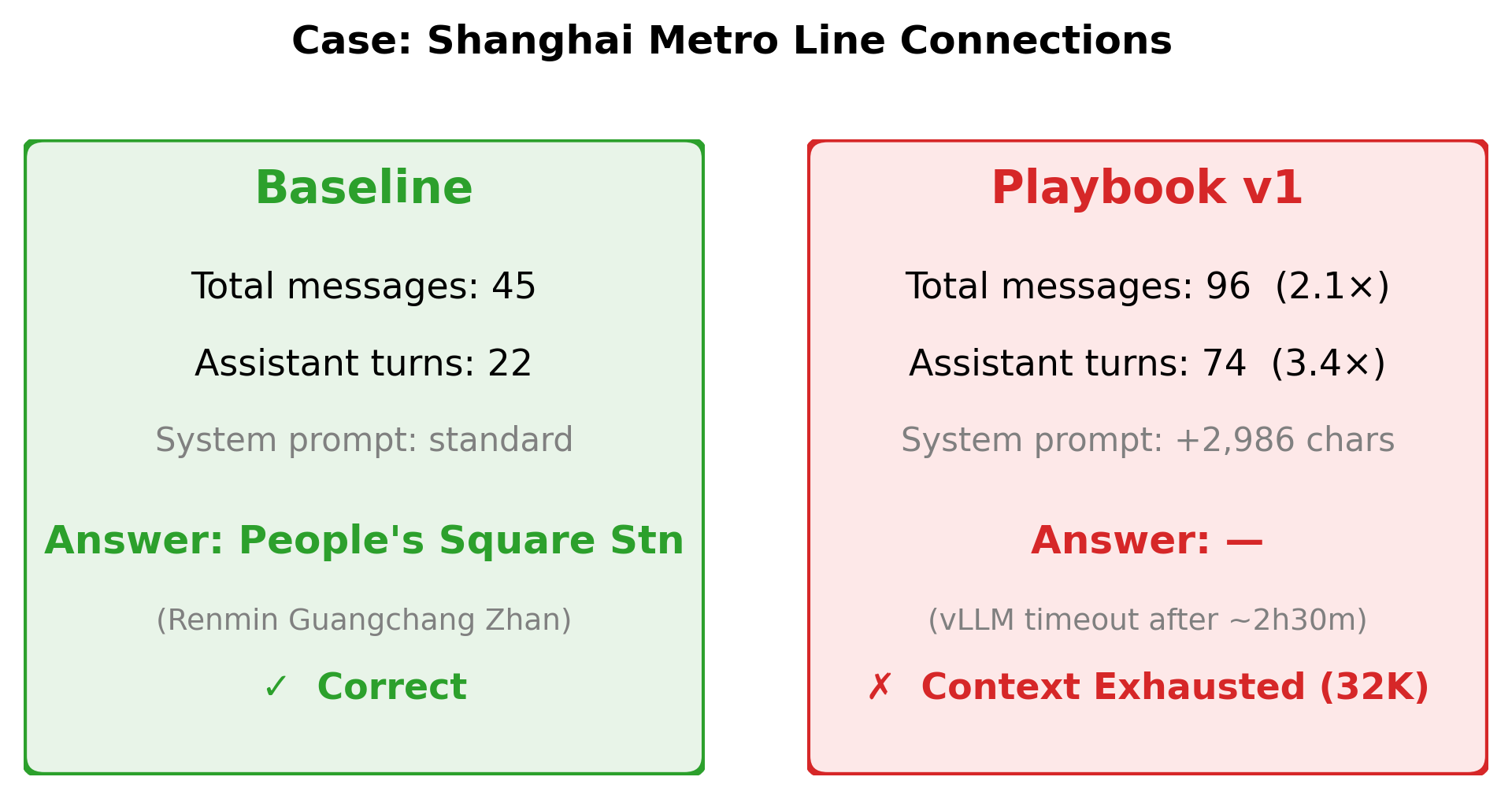}
\caption{Illustrative xbench failure case: Shanghai Metro line connectivity. The question asks which station can be reached from any other station with at most one transfer (answer: People's Square). Baseline solves it correctly in 45 messages / 22 turns; the transferred playbook inflates the trajectory to 96 messages / 74 turns (+2{,}986 chars of system prompt) and terminates without an answer after exhausting the 32K context budget (${\sim}$2.5\,h runtime).}
\label{fig:xbench_case_card}
\end{figure}

\paragraph{Shanghai Metro connectivity.} The question asks which Shanghai Metro station can be reached from any other station with at most one transfer; the correct answer is People's Square. In baseline, the agent answers correctly after 45 total messages and 22 assistant turns. With the transferred playbook, the system prompt grows by 2986 characters, the trajectory expands to 96 messages and 74 assistant turns, and the run terminates without an answer after roughly 2.5 hours. The failure is therefore best understood as \emph{trajectory inflation under a fixed context budget}.

\end{document}